\documentclass[preprint]{article} 

\usepackage{graphicx}
\usepackage{amsmath}
\usepackage{amssymb}
\usepackage{subfigure}
\usepackage{latexsym}

\usepackage[algo2e,ruled,vlined]{algorithm2e}

\graphicspath{
  {fig/eps/}
  {fig/pdf/}
  {.}
}

\newcommand{\orbitone}[1]{\left\langle#1\right\rangle}
\newcommand{\orbit}[2]{\orbitone{#1}\left(#2\right)}
\newcommand{\ma}[1]{\alpha_{#1}} 
\newcommand{\gmap}{$n$-Gmap} 
\newcommand{\gmaps}{$n$-Gmaps} 

\newcommand{\Reffig}[1]{Figure~\ref{#1}} 
\newcommand{\reffig}[1]{Fig.~\ref{#1}} 

\newcommand{\Refsec}[1]{Section~\ref{#1}} 
\newcommand{\refsec}[1]{Sect.~\ref{#1}} 

\newcommand{\reftab}[1]{Table~\ref{#1}} 

\newcommand{\refalgo}[1]{Algo.~\ref{#1}} 

\newcommand{\refdef}[1]{Def.~\ref{#1}} 

\newcommand{\Refprop}[1]{Proposition~\ref{#1}} 
\newcommand{\refprop}[1]{Prop.~\ref{#1}} 

\newtheorem{definition}{Definition}

\newtheorem{proposition}{Proposition}

\newcommand{\proof}{\noindent{\bf Proof. }}

\newcommand{\qed}{\hfill$\Box$}

\title{Removal and Contraction Operations in $n$D Generalized Maps
       for Efficient Homology Computation}

\author{\Large RESEARCH REPORT\\\\Guillaume Damiand\\
\small Universit\'{e} de Lyon, CNRS, LIRIS, UMR5205, F-69622 France
\\\small guillaume.damiand@liris.cnrs.fr\\\\
Rocio Gonzalez-Diaz\\\small
Universidad de Sevilla, Dpto. de Matem\'atica Aplicada I, S-41012, Spain\\\small
rogodi@us.es\\\\
Samuel Peltier\\\small
Universit\'{e} de Poitiers, CNRS, XLIM-SIC, UMR6172, F-86962 France\\\small
samuel.peltier@xlim.fr}

\date{}
\begin{document}
\maketitle
\begin{abstract} 
  In this paper, we show that contraction operations preserve the
  homology of $n$D generalized maps, under some conditions. This
  result extends the similar one given for removal operations
  in~\cite{DGP2012}. Removal and contraction operations are used to
  propose an efficient algorithm that compute homology generators of
  $n$D generalized maps. Its principle consists in simplifying a
  generalized map as much as possible by using removal and contraction
  operations. We obtain a generalized map having the same homology
  than the initial one, while the number of cells decreased
  significantly.

\end{abstract}

\section{Introduction}
\label{sec:intro}

In different areas of computer science, objects are represented as
cells and incidence relations. Most of the time, simplicial or cubical
complexes are used~\cite{May67,Mun84,Hat02,KMM04}.  Then, it is often
required for some high level operations to compute features on the
described objects. These features could be geometric, such as a
curvature estimator, colorimetric, such as an histogram of colors, or
topological, such as Betti numbers. Among the existing topological
features, the computation of homology over different combinatorial
structures has been mainly studied
\cite{KMM04,EdelsbrunnerLZ02,NSKPMT02,basak10,pawel,morozov}.

Most of the time, the representation of a subdivided object using
simplicial or cubical structures require more cells than using a
cellular one, where cells can be more general.
Indeed, when an operation modify these models, it is often required to
apply a post-processing step in order to keep the model valid, for
example a remeshing step for triangle data structures.

To solve these drawback, \gmaps{} have been introduced in
\cite{Lienhardt91,Lienhardt94}. This model allows to describe any
cellular quasi manifolds orientable or not in any dimension. One main
interest of \gmaps{} is to be able to describe cells more general than
only simplicial or cubical cells. This simplify and improve the
efficiency of operations on this model which could be defined locally.
For this reason, \gmaps{} and some variants were used in several
previous works on image processing and geometrical modeling.

Now we are studying the problem of computing features on \gmaps{}, and
particularly on the computation of homology generators. To reach this
objective, a boundary operator has been defined in
\cite{AlayranguesAl09}, and it has been proven in
\cite{AlayranguesAl11} that there is a subclass of \gmaps{} for which
the homology obtained by this operator is equivalent to the homology
of the corresponding simplicial complex.



In this paper, we focus on optimization for computing efficiently
homology generators for this subclass of \gmaps{}.  As the complexity
of homology computation is directly linked to the number of cells of
an object, the optimization focuses on two simplification operations:
removal and contraction of cells. In this paper, we prove that these
operations preserve the homology of a \gmap{}. More precisely, these
two operations allow to obtain an homologous object with few number of
cells. Then we can compute the homology generators on the reduced
object by reducing incidence matrices into their Smith-Agoston normal
form~\cite{Ago76,PAFL06,DHSW03,Storjohann}. We show some experiments
that illustrate the interest of our simplification method when we
compute 2D and 3D homology generators of triangular and cubical
complexes.  Moreover, we are able to directly project the homology
generators computed on the reduced object on the original object.

\Refsec{sec:recalls} recalls all the related materials regarding
\gmaps{}, removal operations and
homology. \Refsec{sec:preserving-contraction} presents the main
results which state that homology is preserved for removal and
contraction operations under some conditions. We present in
\refsec{sec:simplification-algorithm} a simplification algorithm based
of these operations which ensures to preserve the homology of the
described object.  In \refsec{sec:expe} we present some experiments
showing that the number of cells is widely reduced. Finally,
\refsec{sec:conclu} concludes this work and gives some possible
improvements.

\section{Preliminary Works}
\label{sec:recalls}

Generalized maps are combinatorial structures allowing to describe
cellular subdivided objects. They are defined in any dimension, based
on a unique basic elements, called \emph{darts}. The notions of cells,
adjacency and incidence are implicitly encoded though the notion of
orbits and involutions.

\subsection{$n$-Gmaps and Cells}

%
\def\largFig{.3\textwidth}%
\begin{figure}
  \begin{center}
    \subfigure[\label{fig-ngmap-example-a}]
    {\includegraphics[width=\largFig]{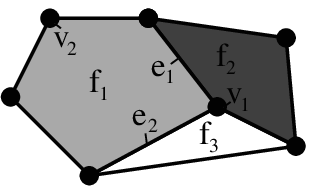}}\qquad
    \subfigure[\label{fig-ngmap-example-b}]
    {\includegraphics[width=\largFig]{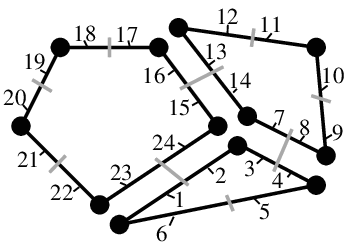}}
    \caption{Example of a 2-Gmap.  \subref{fig-ngmap-example-a}~A 2D
      cellular complex containing 3 faces; 9 edges and 7 vertices.
      \subref{fig-ngmap-example-b}~The 2-Gmap
      $G=(D,\ma{0},\ma{1},\ma{2})$ describing this cellular complex,
      having 24 darts (represented by numbered black segments).  Two
      darts linked by $\ma{0}$ are drawn consecutively and separated
      by a gray segment (for example $\ma{0}(19)=20$), two darts
      linked by $\ma{1}$ share a common point (for example
      $\ma{1}(20)=21$), and two darts linked by $\ma{2}$ are drawn
      parallel, the gray segment over these two darts (for example
      $\ma{2}(13)=16$). }
    \label{fig-ngmap-example}
  \end{center}
\end{figure} 
%
Let us consider the 2D object shown in \reffig{fig-ngmap-example-a} to
give the intuition of what is a generalized map. This object is
composed by 3 faces (2D elements), 9 edges (1D elements) and 7
vertices (0D elements).  This object is described with the
2-dimensional generalized map shown in
\reffig{fig-ngmap-example-b}.
        Intuitively, we decompose each face of
the 2D object in isolated faces, then we decompose each edge of the
isolated faces in isolated edges, and lastly we decompose each
isolated edge in isolated vertices. Elements obtained by this process are called
\emph{darts} and are the atomic basic elements of any generalized map
(numbered segments in \reffig{fig-ngmap-example-b}).  Then we add
relations between these darts to represent the relations broken during
the decomposition process. $\alpha_0$ links two darts that belonged to
the same edge and face before the vertex decomposition (for example
$\ma{0}(19)=20$ in \reffig{fig-ngmap-example-b}); $\alpha_1$ links two
darts that belonged to the same vertex and face before the edge
decomposition (for example $\ma{1}(20)=21$) and $\alpha_2$ links two
darts that belonged to the same vertex and edge before the face
decomposition (for example $\ma{2}(13)=16$).  

The same principle of decomposition can be done in any dimension which
gives the generic definition of $n$-dimensional generalized maps in
\refdef{def-ngmap} \cite{Lienhardt91,Lienhardt94}.
\begin{definition}[$n$-Gmap]\label{def-ngmap}
An \emph{\emph{n}-dimensional generalized map}, called
\emph{\gmap{}}, with $0\leq n$, is a $(n+2)-$tuple 
$G=(D,\ma{0},\ldots,\ma{n})$ 
where:
\begin{enumerate}
\item $D$ is a finite set of darts;
\item $\forall i, \, 0 \leq i \leq n$, $\ma{i}$ is an
  involution on $D$;
\item $\forall i:\, 0 \leq i \leq n-2$,
  $\forall j:\, i+2 \leq j \leq n$, 
  $\ma{i}\circ\ma{j}$ is an involution.
\end{enumerate}
\end{definition}

We retrieve the set of darts $D$, and the $n+1$ relations between
these darts, $\ma{0},\ldots,\ma{n}$. These relations are
\emph{involutions}, i.e. bijection equal to their inverse, because
when two darts are linked by $\ma{i}$, they are linked in both
direction: we have $\ma{i}(d_1)=d_2$ and $\ma{i}(d_2)=d_1$.  Besides,
we say that $d$ is $i-$\emph{free} if $\alpha_i(d)=d$.  Intuitively,
that means that there is no other $i$-cell around dart $d$. In the
example of \reffig{fig-ngmap-example}, darts $5,6,9-12,17-22$ are
$2-$free.  The last line of this definition ensures the topological
validity of the described objects. Intuitively, this condition ensures
that when two darts of two cells are linked, then all the darts of the
cells are two-by-two linked. This ensures that two cells are either
disjointed, or completely linked, but they cannot be partially shared.
For example in 2D, this condition ensures that $\ma{0}\circ\ma{2}$ is
an involution, i.e. in the example of \reffig{fig-ngmap-example-b},
since $\ma{0}(1)=2$, if $\ma{2}(1)=23$ then it is required that
$\ma{2}(2)=24$.

An $n$-Gmap allows to represent all the cells of a subdivided objects and
all the incidence and adjacency relations, thanks to the orbit notion.
Intuitively, given a set of involutions $\Phi$, the orbit of an
element $d$ relatively to $\Phi$ is the set of all the elements that
can be obtained from $d$ by using any combination of any involutions
in $\Phi$.
%

\begin{definition}[Orbit]\label{def-orbit} 
  Let $\Phi=\{\pi_0,\cdots,\pi_n\}$ be a set of involutions defined
  on a set $D$.  $\orbitone{\Phi}$ is the involution group of $D$
  generated by $\Phi$. The \emph{orbit} of an element $d\in D$
  relatively to $\orbitone{\Phi}$, denoted $\orbit{\Phi}{d}$ is the
  set $\{\phi(d) \mid \phi \in \orbitone{\Phi}\}$.
\end{definition}

The cells of an $n$-Gmap are defined by some specific orbits.
%
\begin{definition}[$i$-cell]\label{def-cell} 
  Let $G$ be an $n$-Gmap, and $d \in D$ be a dart.  Given $i$, $0\leq
  i \leq n$, the $i$-dimensional cell containing $d$, called
  \emph{$i$-cell} and denoted by $c^i(d)$, is
  $\orbit{\ma{0},\ldots,\ma{(i-1)},\ma{(i+1)},\ldots,\ma{n}}{d}$.
\end{definition}

Intuitively, as $\ma{i}(d)$ gives the dart belonging to another
$i$-cell than the $i$-cell containing dart $d$, considering the orbit
containing all the involutions of the $n$-Gmap except $\ma{i}$ gives
all the darts that belong to the same $i$-cell than $d$: in the
generalized map framework, this set of darts is the $i$-cell.

Observe that if a dart $e$ belongs to an $i$-cell $c^i(d)$ then,
$c^i(e)=c^i(d)$. Besides, each dart belongs to exactly one cell in
each dimension. Therefore, each cell $c$ can be uniquely given by a
set of darts and its dimension. Given an $n$-Gmap $G$, $S_G^i$ denotes
the set of all the cells (set of darts) of dimension $i$ and
$S_G=\{S_G^q\}_q$ is the graded set of all the cells obtained from the
$n$-Gmap $G$.

Two $i$-cells $c_1$ and $c_2$ are \emph{adjacent} if there is two
darts $d_1\in c_1 $ and $d_2\in c_2$ such that $\ma{i}(d_1)=d_2$. Two
cells $c_3$ and $c_4$ are \emph{incident} if $c_3 \neq c_4$ and if
$c_3 \cap c_4 \neq \emptyset$.

In the example of \reffig{fig-ngmap-example}, face $f_3$ is described
by $\orbit{\ma{0},\ma{1}}{1}=\{1,2,3,4,5,6\}$, edge $e_1$ 
by $\orbit{\ma{0},\ma{2}}{13}=\{13,14,15,16\}$, and vertex $v_1$ 
by $\orbit{\ma{1},\ma{2}}{2}$ $=\{2,3,7,14,152,24\}$. $v_1$ and
$e_1$ are incident since $\orbit{\ma{1},\ma{2}}{2}$ 
$\cap \orbit{\ma{0},\ma{2}}{13}$ $=\{14,15\}\neq \emptyset$.  $f_1$ and $f_3$
are adjacent since $23 \in f_1$, $1 \in f_3$, and $\ma{2}(1)=23$.

\subsection{Removal and Contraction Operations}

Now, we want to simplify a given $n$-Gmap by deacreasing its number of
cells. For that, we are going to use two basic operations: the
\emph{removal} and the \emph{contraction} of a cell
\cite{DamLie03}. Firstly, we introduce the removal operation which
consists to remove an $i-$cell, while merging its two incident
$(i+1)-$cells. This operation is not always possible: the cell to
remove must be removable.  The contraction operation can be
defined in a similar way than the removal operation. Indeed, these two
operations are dual: removing an $i-$cell in an $n$-Gmap is equivalent
to contracting the corresponding $(n-i)-$cell in the dual $n$-Gmap.

%
\begin{definition}[Removable and contractible cells]\label{def-cellule-supprimable}
  Let $G$ be an $n$-Gmap and $c$ an $i$-cell of $G$. 
  \begin{itemize}
  \item $c$ is \emph{removable}\\ if $i=n-1$, or if $0\leq i<n-1$ and
    $\forall d \in c,
    \ma{i+1}\circ\ma{i+2}(d)=\ma{i+2}\circ\ma{i+1}(d)$.
  \item $c$ is \emph{contractible}\\if $i=1$, or if $1<i\leq n$ and
    $\forall d \in c,
    \ma{i-1}\circ\ma{i-2}(d)=\ma{i-2}\circ\ma{i-1}(d)$.
  \end{itemize}
\end{definition}
%

The notion of removable cell $c$ is strongly related to the number of
its $(i+1)$ incident cells, called the {\it degree} of $c$ and denoted
$degree(c)$.  Similarly, the notion of contractible cell $c$ is
strongly related to the number of its $(i-1)$ incident cells, called
the {\it codegree} of $c$ and denoted $codegree(c)$.  A consequence of
\refdef{def-cellule-supprimable} is that an $i-$cell of degree $> 2$
is not removable and an $i-$cell of codegree $> 2$ is not
contractible.
%

Now we can define the $i-$removal operation. This operation takes an
$n$-Gmap and an $i$-cell $c$ to remove as input, and modify the
$n$-Gmap to obtain the generalized map in which $c$ is removed.
%
\begin{definition}[$i$-removal]\label{def-suppression}
  Let $G=(D,\alpha_0,\dots,\alpha_n)$ be an $n$-Gmap and $c$ be a removable $i$-cell of $G$.  We denote
  $DV = \ma{i}(c) \setminus c$, the set of darts $i$-linked with $c$
  that do not belong to $c$. The $n$-Gmap obtained by removing $c$ from
  $G$ is $G'=(D',\ma{0}',\dots,\ma{n}')$ defined by:
  \begin{itemize}
  \item $D' = D \setminus c$;
  \item $\forall j \in \{0,\ldots,i-1,i+1,\ldots,n\}:
    \ma{j}' = \ma{j}|D'$; 
    \footnote{$\ma{j}'$ is equal to $\ma{j}$ restricted to
      $D'$, \emph{i.e.} $\forall d \in D': \ma{j}'(d)= \ma{j}(d)$.}
  \item $\forall d \in D' \setminus DV: \ma{i}'(d) = \ma{i}(d)$;
  \item $\forall d \in DV: \ma{i}'(d)=\left(
      \ma{i}\circ \ma{i+1} \right)^{k} \circ \ma{i}(d)$,\\
    with $k$ the smallest integer such that $\left(
      \ma{i}\circ \ma{i+1} \right)^{k}\circ  \ma{i}(d) \in DV$.
  \end{itemize}
\end{definition}

In the $2$-Gmap shown in \reffig{fig-removal-example-d}, which describes
the 2D subdivided object shown in \reffig{fig-removal-example-a}, all
the edges are removable (since an $(n-1)$-cell is always removable in
an $n$-Gmap), vertex $v_2$ is removable while vertex $v_1$ is
not. Removing edge $e_1$ merges faces $f_1$ and $f_2$ in one face,
called $f'_1$, having as boundary the boundary of $f_1$ plus the
boundary of $f_2$ minus edge $e_1$. We obtain the 2-Gmap shown in
\reffig{fig-removal-example-e} which corresponds to the subdivided
object shown in \reffig{fig-removal-example-b}.  In this $2$-Gmap,
vertex $v_3$ is now removable (while it was not removable before the
removal of edge $e_1$), and we remove it. Its two incident edges,
$e_3$ and $e_4$, are merged in one edge, called $e'_3$.  We obtain the
$2$-Gmap shown in \reffig{fig-removal-example-f} which corresponds to
the subdivided object shown in \reffig{fig-removal-example-c}.
%
\def\largFig{.3\textwidth}%
\begin{figure}
  \begin{center}
    \subfigure[\label{fig-removal-example-d}]
    {\includegraphics[width=\largFig]{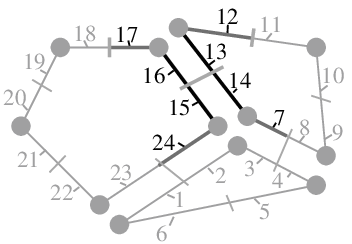}}\quad
    \subfigure[\label{fig-removal-example-e}]
    {\includegraphics[width=\largFig]{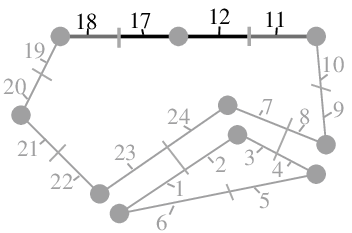}}\quad
    \subfigure[\label{fig-removal-example-f}]
    {\includegraphics[width=\largFig]{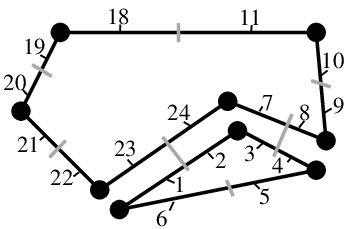}}\\
    \subfigure[\label{fig-removal-example-a}]
    {\includegraphics[width=\largFig]{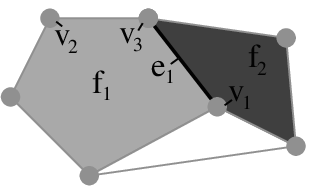}}\quad
    \subfigure[\label{fig-removal-example-b}]
    {\includegraphics[width=\largFig]{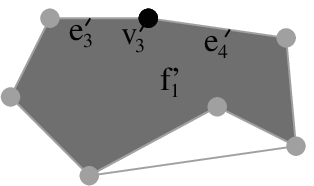}}\quad
    \subfigure[\label{fig-removal-example-c}]
    {\includegraphics[width=\largFig]{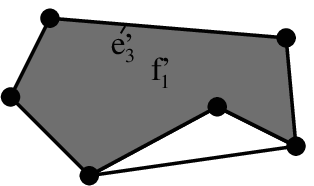}}
    \caption{Examples of removal operations in $2$-Gmaps, and the
      corresponding effect on the described objects. The first line
      gives the $2$-Gmaps, and the second line the corresponding 2D
      subdivided objects. \subref{fig-removal-example-d} and
      \subref{fig-removal-example-a}: Initial configuration.
      \subref{fig-removal-example-b} and
      \subref{fig-removal-example-e}: Configuration obtained from the
      initial configuration by removing edge $e_1$, described by darts
      $\{13,14,15,16\}$.
      \subref{fig-removal-example-c} and
      \subref{fig-removal-example-f}: Configuration obtained from the
      second configuration by removing vertex $v_3$, described by darts
      $\{12,17\}$.      
    }
    \label{fig-removable-example}
  \end{center}
\end{figure}

\begin{definition}[$i$-contraction]\label{def-contraction}
  Let $G=(D,\alpha_0,\dots,\alpha_n)$ be an $n$-Gmap and $c$ be a
  contractible $i$-cell of $G$.  We denote $DV = \alpha_i(c)\setminus
  c$, the set of darts $i$-linked with $c$ that do not belong to
  $c$. The $n$-Gmap obtained by contracting $c$ from $G$ is
  $G'=(D',\ma{0}',\dots,\ma{n}')$ defined by:
\begin{itemize}
\item $D' = D \setminus c$;
\item $\forall j \in  \{0,\ldots,i-1,i+1,\ldots,n\}$: $\alpha'_j = \alpha_j
  |_{D'}$;
\item $\forall d \in D' \setminus DV$: $\alpha'_i(d)=\alpha_i(d)$;
\item $\forall d \in DV$:
  $\alpha'_i(d)=(\alpha_i\circ \alpha_{i-1})^{k}\circ \alpha_i(d)$,\\
  with $k$ the smallest integer such that
  $(\alpha_i\circ \alpha_{i-1})^{k}\circ \alpha_i(d) \in DV$.
\end{itemize}
\end{definition}

Example of contractible cells and contraction operations are given in
\reffig{fig-contractible-example}.

The $n$-Gmap obtained by removing/contracting $c$ from $G$ is $G'$,
where we have removed all the darts of $c$ from its set of darts;
where all the involutions $\ma{j}$ for $j\neq i$ are preserved; where
$\ma{i}(d)$ is preserved for each dart $d$ that is not $i$-linked to
one dart of $c$.  Thus the only modification concern $\ma{i}(d)$ for
each dart $d$ which is $i$-linked to one dart of $c$.  For such a
dart, we modify its $\ma{i}$ to be the first dart found after
traversing darts of $c$.  The only difference between removal and
contraction operations is the way that we traverse darts of $c$: we
use successively $\alpha_i\circ \alpha_{i+1}$ for removal, while we
use successively $\alpha_i\circ \alpha_{i-1}$ for contraction.


\def\HautFig{2.2cm}
\begin{figure}
  \begin{center}
    \subfigure[\label{fig-contr1}]
    {\includegraphics[height=\HautFig]{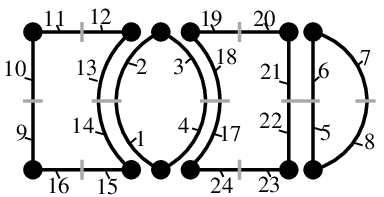}}\quad
    \subfigure[\label{fig-contr2}]
    {\includegraphics[height=\HautFig]{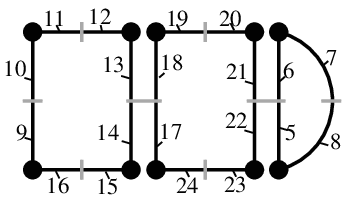}}\quad
    \subfigure[\label{fig-contr3}]
    {\includegraphics[height=\HautFig]{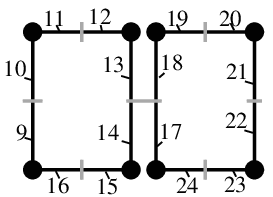}}
   \caption{Examples of contraction operations in 2-Gmaps.
     \subref{fig-contr1}  Initial configuration where the two faces
     $\{1,2,3,4\}$ and $\{5,6,7,8\}$ are contractible.     
     \subref{fig-contr2} 2-Gmap obtain from the initial configuration
     by contracting the face $\{1,2,3,4\}$.
      \subref{fig-contr3} 2-Gmap obtain from the second configuration
      by contracting the face $\{5,6,7,8\}$.}
    \label{fig-contractible-example}
  \end{center}
\end{figure}

\subsection{Introduction to Homology }

Homology is a topological invariant that characterizes $k-$dimensional
holes of an object (i.e. connected components, tunnels, cavities...).
Homology groups are defined from an algebraic structure called free
chain complex, denoted $(C_*, \partial_*)$. Each group $C_k$ is the
group of $k-$chains, generated by all the $k-$cells. The homomorphisms
$\partial_k$ describe the boundary of $k-$chains as $(k-1)-$chains. In
particular, the boundary of any $0-$chain is trivial, and for any
$k-$chain, $k>1$, $\partial_{k-1}\circ\partial_{k}=0$.  Homology can
be computed over different coefficient group ($\mathbb{Z}, \mathbb{Z}
/2\mathbb{Z}, \mathbb{Q}...)$, but the most topological information is
obtained when computing homology over $\mathbb{Z}$. Thus, for computing
homology, we need a boundary operator. 

\subsection{Homology for $n$-Gmaps}

Now the question is how to compute the homology of $n$-Gmaps.  For
that, we have defined a boundary operator
\cite{AlayranguesAl09,AlayranguesAl11} for \gmaps{}.  However, this
boundary operator is defined only for orientable cells.

In an $n$-Gmap, a cell is orientable if it can be partitioned in two
subsets of darts, such that two darts linked by any $\ma{j}$ do not
belong to the same subset. Note that vertices are always orientable.
%
\begin{definition}[Orientable $i$-cell]\label{def-orientable-cell}
  An $i$-cell $c$ is \emph{orientable} if $i=0$ or if $c=e_1\cup{}e_2$
  such that: $\forall d \in c$, $\forall j$, $0\leq j\leq n$, $j\neq
  i$: $d$ is not $j$-free $\Rightarrow$ $d$ and $\ma{j}(d)$ do not
  belong to the same set $e_1$ or $e_2$.  $c$ is \emph{non-orientable}
  otherwise.
\end{definition}

Note that a non orientable object can have all its cells orientable.
For example, the $2$-Gmap in \reffig{fig-ngmap-orientable-cell-a}
represents a M\"obius strip, which is non orientable object, but all
its cells are orientable.  The second example, presented in
\reffig{fig-ngmap-orientable-cell-b}, describes a $3$-Gmap having a
non orientable $3$-cell.
\def\largFig{.3\textwidth}%
\begin{figure}
  \begin{center}
    \subfigure[\label{fig-ngmap-orientable-cell-a}]
    {\includegraphics[width=\largFig]{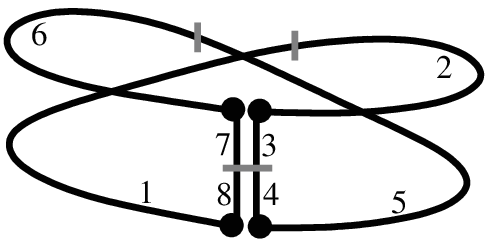}}\qquad
    \subfigure[\label{fig-ngmap-orientable-cell-b}]
    {\includegraphics[width=\largFig]{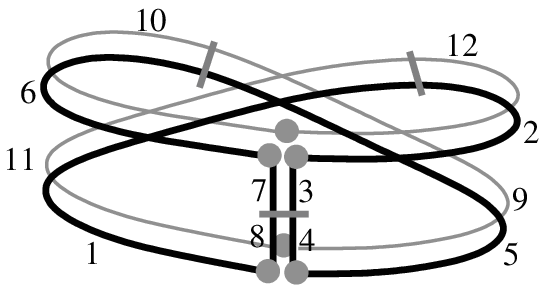}}\\
      \subfigure[\label{fig-ngmap-orientable-cell-c}]
      {$\begin{array}{c|cccccccccccc}
          &1&2&3&4&5&6&7&8&9&10&11&12\\
          \hline
          \alpha_0&2&1&4&3&6&5&8&7&10&9&12&11\\
          \alpha_1&8&3&2&5&4&7&6&1&11&12&9&10\\
          \alpha_2&11&12&7&8&9&10&3&4&5&6&1&2
        \end{array}$\newline}
      \caption{Example of orientable and non-orientable
        cells. \subref{fig-ngmap-orientable-cell-a}~A $2$-Gmap containing
        $8$ darts, and representing a M\"obius strip. All the cells of this
        $2$-Gmap are orientable, even the face $\{1,2,3,4,5,6,7,8\}$, but
        the $2$-Gmap itself is non orientable.
        \subref{fig-ngmap-orientable-cell-b}~A $3$-Gmap containing $12$
        darts, and representing a volume which boundary is a projective
        plane, i.e. the closure of a M\"obius strip by adding a new face. In
        this $3$-Gmap, all the $0$-cells, $1$-cells and $2$-cells are
        orientable. But the $3$-cell $\{1,2,3,4,5,6,7,8,9,10,11,12\}$ is not
        orientable.  \subref{fig-ngmap-orientable-cell-c}~Involutions
        $\alpha_0,\alpha_1,\alpha_2$ of the $3$-Gmap represented in (b).
        $\alpha_3=id$ for each dart.}
      \label{fig-ngmap-orientable}
    \end{center}
\end{figure} 

Now given an orientable $i$-cell, we have to orient it. As cells are
described by set of darts in $n$-Gmaps, the orientation of a cell
will be made through the orientation of its darts. We associate to
each dart $d$ a sign, denoted $sg^i(d)$, that gives the orientation of
dart $d$ for its $i$-cell.
%
\begin{definition}[Signed $i$-cell]\label{def-signed-cell} 
  Let $c$ be an orientable $i$-cell.  The corresponding {signed
    $i$-cell} is $c$ together with a sign for each of its dart $d$, denoted
  $sg^i(d)$:
  \begin{itemize}
  \item[$\bullet$] $sg^i(d)=-sg^i(\ma{j}(d))$ $\forall j$: $0 \leq j < i$ such
    that $d$ is not $j$-free; 
  \item[$\bullet$] $sg^i(d)=sg^i(\ma{j}(d))$ $\forall j$: $i< j \leq n$.
  \end{itemize}
\end{definition}

We can see in \reffig{fig-signed-cells-example} the signed cells for
the 2-Gmap introduced in
\reffig{fig-ngmap-example}. \Reffig{fig-signed-cells-example-d} shows
$sg^0$ and the orientation of 0-cells,
\reffig{fig-signed-cells-example-e} shows $sg^1$ and the orientation
of 1-cells, and \reffig{fig-signed-cells-example-f} shows $sg^2$ and
the orientation of 2-cells.  The corresponding orientation of
$i$-cells are shown on the second line of the figure, above the
$2$-Gmap with the corresponding signed incidence numbers $sg^i$. Note
that the choice of the orientation of each cell is totally arbitrary,
and has no consequence on the homology computation.

%
\def\largFig{.3\textwidth}%
\begin{figure}
  \begin{center}
    \subfigure[\label{fig-signed-cells-example-d}]
    {\includegraphics[width=\largFig]{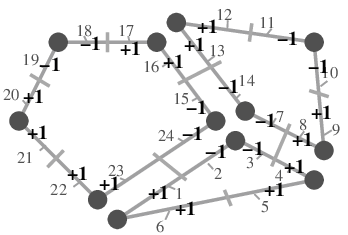}}\quad
    \subfigure[\label{fig-signed-cells-example-e}]
    {\includegraphics[width=\largFig]{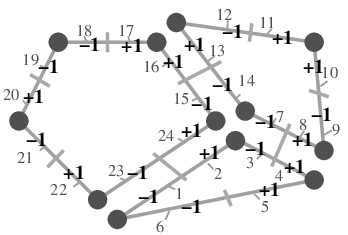}}\quad
    \subfigure[\label{fig-signed-cells-example-f}]
    {\includegraphics[width=\largFig]{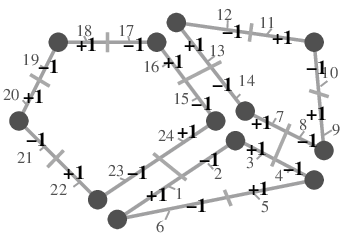}}\\
    \subfigure[\label{fig-signed-cells-example-a}]
    {\includegraphics[width=\largFig]{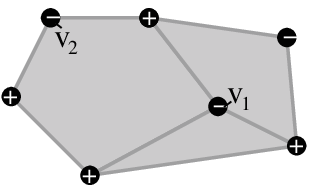}}\quad
    \subfigure[\label{fig-signed-cells-example-b}]
    {\includegraphics[width=\largFig]{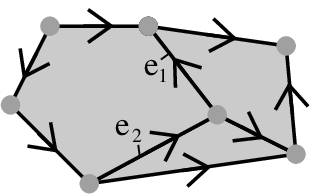}}\quad
    \subfigure[\label{fig-signed-cells-example-c}]
    {\includegraphics[width=\largFig]{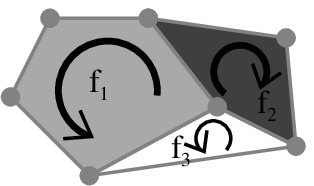}}
    \caption{Examples of signed incidence number in 2-Gmaps, and the
      corresponding orientation on the cells of the subdvided
      objects. The first line gives the 2-Gmaps, and the second line
      the corresponding 2D subdivided objects with the oriented cells.
      \subref{fig-signed-cells-example-d} and
      \subref{fig-signed-cells-example-a}~Orientations of 0-cells.
      \subref{fig-signed-cells-example-e} and
      \subref{fig-signed-cells-example-b}~Orientations of 1-cells.
      \subref{fig-signed-cells-example-f} and
      \subref{fig-signed-cells-example-c}~Orientations of 2-cells.}
    \label{fig-signed-cells-example}
  \end{center}
\end{figure}

As we can see in \reffig{fig-signed-cells-example-d}, all the darts of
a same $0$-cell have the same sign $sg^0$.  For 1-cells, two darts of
the same $1$-cell have the same sign $sg^1$ if they are linked with
$\ma{2}$, and they have two opposite signs if they are linked with
$\ma{0}$. In \reffig{fig-signed-cells-example-e}, we have for example
$\ma{0}(13)=14$ and $\ma{2}(13)=16$. Thus darts $13$ and $14$ have two
opposite signs $sg^1(13)=+1$ and $sg^1(14)=-1$; and darts $13$ and
$16$ have the same sign $sg^1(13)=+1$ and $sg^1(16)=+1$.  In
\reffig{fig-signed-cells-example-b}, we choose the convention that a
$1-$cell is oriented starting from its $-1$ extremity and going to its
$+1$ extremity, but we can consider the other convention used here
only on the figure. For $2$-cells, two darts of the same $2$-cell have
two opposite signs $sg^2$ if they are linked with $\ma{0}$ or with
$\ma{1}$ (see \reffig{fig-signed-cells-example-f}). In
\reffig{fig-signed-cells-example-c} we choose the convention that a
2-cell is oriented by turning starting from a negative dart and going
to a positive one.

%
\begin{definition}[Signed incidence number]
\label{def:signed-inc-num}
Let $G$ be an $n$-Gmap with all its cells signed. Let $c^i$ be an
$i$-cell of $G$ and $d$ one of its darts.  Let $\{p_j\}_{j=1 \cdots
  k}$ be a set of darts such that the orbits
$\{\orbit{\ma{0},\cdots,\ma{(i-2)}}{p_j}\}_{j=1 \cdots k}$ make a
partition of $\orbit{\ma{0},\ldots,\ma{(i-1)}}{d}$.  The \emph{signed
  incidence number} between the cell $c^i$ and an $(i-1)$-cell
$c^{i-1}$ is defined by
$$(c^i:c^{i-1})=\sum\limits_{p_j, j=1 \cdots k | p_j \in c^{i-1}} sg^i(p_j).sg^{i-1}(p_j).$$
\end{definition}

Consider the $2$-Gmap in \reffig{fig-signed-cells-example}.  Let
$i=1$, $d=15$ and $c^1=e_1=\{13,14,15,16\}$.  Then,
$\orbit{\ma{0}}{15}=\{15,16\}$; $k=2$; $p_1=15$ and $p_2=16$. Let
$c^0=v_1=\{2,3,7,14,15,24\}$. Then,
$$(e_1:v_1)=\sum\limits_{p_j,j=1,2| p_j \in v_1} sg^1(p_j).sg^{0}(p_j)= sg^1(15).sg^{0}(15)=1.$$

Now the boundary operator $\partial_{G}$ of any $i$-cell $c$ is
defined as $\partial_G(c)=\sum_{c'}(c:c')c'$, where $c'$ are
$(i-1)-$cells incident to $c$.  The boundary operator $\partial_{G}$
satisfies $\partial_{G}\circ \partial_{G} = 0$ when there are no
$i$-free darts for $0 \leq i \leq n-1$.  Observe that if there are no
$i$-free darts for $0 \leq i \leq n-1$, neither are after removals and
contractions.

Let $C_q(S_G)$ denote the group of $q$-chains of $S_G$.  The
\emph{chain complex} $(C_*(S_G),$ $\partial_G)$ is the chain group
$C_*(S_G)=\{C_q(S_G)\}_q$ together with the boundary operator
$\partial_G$.  The homology of $G$ is defined as the homology of the
chain complex $(C_*(S_G),\partial_G)$.

We have proven in \cite{AlayranguesAl11} that the homology
defined on \gmaps{} by this boundary operator is equivalent to the
simplicial homology of the associated quasi-manifolds when the
homology of the canonical boundary of each $i$-cell is that of an
$(i-1)$-sphere, and when $\forall d \in D$, $\forall i \in
\{0,\ldots,n\}$, $d$ is $i$-free or $\ma{i}(d) \not\in
\orbit{\ma{0},\ldots,\ma{i-2},\ma{i+2},\ldots,\ma{n}}{d}$.  

In the following, all the considered \gmaps{} have all its cells
signed, no $i$-free darts for $0\leq i<n$ and satisfied these
conditions.

\section{Removal and Contraction Operations Preserving Homology}
\label{sec:preserving-contraction}

In this section, we show that under certain conditions, contraction
and removal operations can be performed while preserving the homology.
In particular, we perform removal of degree two cells, and contraction
of codegree two cells.

We start to show that removal and contraction operations preserve the
orientation of each cell. This property is required to guaranty that
we are able to compute the homology of the simplified $n$-Gmap.
\begin{proposition}
  After a removal or contraction of an $n$-Gmap having all its cells
  orientable, we obtain a new $n$-Gmap with all its cell orientable.
\end{proposition}

\proof   We study here the case of degree two cell removal.  We denote by
  $G=(D,\alpha_0,\dots,$ $\alpha_n)$ the initial $n$-Gmap having all
  its cells orientable and $G'=(D',\alpha'_0,\dots,\alpha'_n)$ the map
  obtain after the operation.  Let us prove that after the removal of
  an $i$-cell $c$, cells in $G'$ remain orientable.  $c'$ in $G'$
  orientable means that $c'$ is partitioned in two sets $S_1$ and
  $S_2$ so that two darts $d$ and $d'$ in $c'$ linked by an $\alpha_k$
  belong to two different sets.

  For $j=i$: $c'$ is orientable in $G'$ because $i$-cells
  different from $c$ are not modified by the removal.

  Otherwise $j \neq i$: If $c'$ was a $j$-cell in $G$, for any
  dart $d$, $\alpha_k'(d)=\alpha_k(d)$.  If there is a $j$-cell $e$
  in $G$ such that $c' =e\setminus c$ only $\alpha_i'$ is modified,
  thus for any dart $d$, for any $k\neq i$,
  $\alpha_k'(d)=\alpha_k(d)$.  Thus as $d$ is not $k$-free, $d$ and
  $\alpha_k'(d)$ belongs to two different sets $S_1$ and $S_2$ in
  $G'$ as this is the case for $d$ and $\alpha_k(d)$ in $G$.
  
  We use the same argument for darts $d \in D' \setminus DV$ and for
  $k=i$ as in this case we also have  $\alpha_i'(d)=\alpha_i(d)$.
    
  Now for $d \in DV$ and for $k=i$, we have
  $\alpha_i'(d)=(\alpha_i\alpha_{i+1})^k\alpha_i(d)$.  
  We know that all darts in the path $(\alpha_i\circ
  \alpha_{i+1})^k\circ \alpha_i(d)$ are non free for the next $\alpha$
  used ($d$ is not $i$-free, $\alpha_i(d)$ is not $(i+1)$-free,
  $\alpha_i\circ \alpha_{i+1}(d)$ is not $i$-free \ldots).  As $e$ is
  orientable in $G$, we know that $d$ and $\alpha_i(d)$ belongs to two
  different sets, then $d$ and $\alpha_i\circ \alpha_{i+1}(d)$ belongs
  to the same set \ldots As the length of the part is odd, and no dart
  of the path is free, we conclude that $d$ and $(\alpha_i\circ
  \alpha_{i+1})^k\circ \alpha_i(d)$ belong to two different sets; thus
  $c'$ is orientable in $G'$.

  Otherwise, $j=i+1$ and $c'=a \cup b \setminus c$, with $a$ and $b$
  the two $(i+1)$-cells incident to $c$ in $G$. $a$ and $b$ are
  orientable so there exist two sets that partition each cell: $S'_a$,
  $S''_a$ for $a$ and $S'_b$, $S''_b$ for $b$.  Let us consider
  $S_{a1}$ for the set among $S'_a$, $S''_a$ and $S_{b2}$ for the set
  among $S'_b$, $S''_b$, such that it exists $d \in S_{a1}$ and
  $\alpha_{i+1}(d)\in S_{b2}$; and $S_{a2}$ and $S_{b1}$ for the other
  sets. We know such a dart exists by definition of adjacency
  relation.
  
  Let $S_1=S_{a1} \cup S_{b1}\setminus c$ and $S_2=S_{a2} \cup S_{b2}
  \setminus c$.  Consider two darts $d_1$ and $d_2$ in $c'$ such that
  $\alpha_k'(d_1)=d_2$.  If $k \neq i$, or $d_1 \not\in DV$, we have
  $\alpha_k'(d_1)=\alpha_k(d_1)$ thus $d_1$ and $d_2$ belong to the
  same $(j+1)$-cell $a$ or $b$. Thus we have $d_1 \in S_{a1}$ and $d_2
  \in S_{a2}$ or $d_1 \in S_{b1}$ and $d_2 \in S_{b2}$. Thus $d_1 \in
  S_1$ and $d_2 \in S_2$.
  
  If $d_1 \in DV$ and $k=i$, then $\alpha'_k(d_1)=(\alpha_i\circ
  \alpha_{i+1})^k\circ \alpha_i(d_1)=d_2$. By using the same
  arguments than for above, we conclude that $d_1$ and $d_2$ belong
  to different sets $S_1$ and $S_2$.

  The proof is the same for contraction operation, replacing  $\alpha_{i+1}$
  by $\alpha_{i-1}$.\qed

\Refprop{prop-contraction-degree2} ensures that if an $i-$cell $c$ is
removable and degree two, then $c$ appears $\pm 1$ time in the
boundary of each of its two incident $(i+1)-$cells.  Similarly, if an
$i$-cell is contractible and codegree two, then only its two
$(i-1)-$incident cells appear in the boundary of $c$.

\begin{proposition}\label{prop-contraction-degree2}
  Let $c$ be an $i$-cell, $0\leq i\leq n$.  
  \begin{itemize}
  \item  If $c$ is removable and degree two, then there are two $(i+1)$-cells
    $a$ and $b$ satisfying: $|(a:c)|=|(b:c)|=1$ and for all other
    $(i+1)$-cells $c'$, $(c':c)=0$.
  \item  If $c$ is contractible and codegree two, then there are two
    $(i-1)$-cells $a$ and $b$ satisfying: $|(c:a)|=|(c:b)|=1$ and for
    all other $(i-1)$-cells $c'$, $(c:c')=0$.
  \end{itemize}
\end{proposition}

\proof   Let us consider the case where $c$ is contractible.  Since $c$ is
  codegree two, then there are exactly two $(i-1)-$cells $a$ and $b$
  that are incidence to $c$. Then, for all other $(i-1)$-cells $c'$,
  $(c:c')=0$.  So each dart of $c$ is either in $a$ or in $b$ and
  there exist two darts $d_a, d_b \in c$ such that
  $a=<\alpha_1,...,\alpha_{i-2},\alpha_i,...,\alpha_n>(d_a)$ and
  $b=<\alpha_1,...,\alpha_{i-2},\alpha_i,...,\alpha_n>(d_b)$ and
  $\alpha_{i-1}(d_a)=d_b$.


  Let $d\in s_a$, then $d=\alpha_{i_1}\circ ... \circ
  \alpha_{i_k}(d_a)$. Where $i_1,...,i_k \in
  \{0,\ldots,i-2,i+1,\ldots,n\}$. Similarly, any dart of $s_b$ can be
  written as a composition of $\alpha_k$.  From the definitions of
  Gmaps and contractible cells, then $\alpha_{i-1}\circ
  \alpha_k=\alpha_k\circ \alpha_{i-1}$, so we have $\alpha_{i-1}(d)=
  \alpha_{i-1}\circ \alpha_{i_1}\circ ... \circ \alpha_{i_k}(d_a) =
  \alpha_{i_1}\circ ... \circ \alpha_{i_k}\circ \alpha_{i-1}(d_a) =
  \alpha_{i_1}\circ ... \circ \alpha_{i_k}(d_b) \in s_b$. Then there
  is no dart $d\in s_a$ such that $\alpha_{i-1}(d)\in s_a$ which
  implies that $|(c:a)|=1$.

  The same result holds for darts of $s_b$.  

  The same proof can be done for a removable cell, replacing $i-1$ by
  $i+1$.  \qed

In~\cite{KMS98}, given a chain complex $(C_*(S),\partial)$, it is
proven that if there exist two elements $c\in S_i$ and $c'\in
S_{i+1}$ for some $i>0$ such that $|(c':c)|=1$, then homology is
preserved after removing $c$ from $S_i$ and $c'$ from $S_{i+1}$ and
modifying $\partial$ in a proper way.  Adapting that result to our
purpose, we have the following:

\begin{proposition}\label{prop-general-operation}
  Let $(C_*(S),\partial)$ and $(C_*(S'),\partial')$ be two chain
  complexes.  Let $c$ be an $i$-cell and $c'$ an $(i+1)$-cell, both in
  $S$, such that $|(c':c)|=1$.  Let $\pi:S\setminus \{c,c'\}\to S'$ be
  a bijective function such that for any $j$-cell $x\in
  S\setminus\{c,c'\}$:
\begin{itemize}
\item $\pi(x)$ is a $j$-cell in $S'$;
\item  $\partial'\pi(x)=\pi\left(\partial(x)-(x:c')(c'\right)$ if $j=i+2$;\\
  $\partial'\pi(x)=\pi\left(\partial(x)-(x:c)(c':c)\partial(c')\right)$ if $j=i+1$;\\
  and $\partial'\pi(x)=\pi\partial(x)$ otherwise;
\end{itemize}
where $\pi$ is extended by linearity to chains.  Then the chain
complexes $(C_*(S),\partial)$ and $(C_*(S'),\partial')$ have
isomorphic homology groups.
\end{proposition}

To prove the result, we construct a chain contraction \cite{mclane} of
$(C_*(S),\partial)$ to $(C_*(S'),\partial')$ which is a triple
$(f=\{f_q:C_q(S)\rightarrow C_q(S')\}_q$, $g=\{g_q:C_q(S')\rightarrow
C_q(S)\}_q$ and $\phi=\{\phi_q:C_q(S)\rightarrow C_{q+1}(S)\}_q)$ such
that: (i)~$f$ and $g$ are chain maps;
i.e. $f_{q}\circ\partial_q=\partial'_q\circ{}f_{q}$ and
$g_{q}\circ\partial'_q=\partial_q\circ{}g_{q}$ for all $q$;
(ii)~$\phi$ is a chain homotopy of $id_{C_*\left(
    S\right)}=\{id_q:C_q\left( S\right) \rightarrow C_q\left( S\right)
\}_q$ to $g\circ{}f=\{g_q\circ{}f_q:C_q(S)\to C_q(S)\}_q$; i. e.
$\phi_{q-1}\circ{}\partial_{q}+\partial_{q+1}\circ{}\phi_{q}=id_{q}-g_q\circ{}f_{q}$
for all $q$; (iii)~$f\circ{}g=id_{C_*\left( S'\right) }$.  If a chain
contraction $(f,g,\phi)$ of $(C_*(S),\partial)$ to
$(C_*(S'),\partial')$ exists, then the chain complexes
$(C_*(S),\partial)$ and $(C_*(S'),\partial')$ have isomorphic homology
groups.

\proof   Define $(f,g,\phi)$ as follows: 
\begin{itemize}
\item $f(c)=\pi\left(c-(c':c)\partial(c')\right)$, $f(c')=0$ and
  $f(x)=\pi(x)$ for $x\in S\setminus\{c,c'\}$.
\item Let $z$ be a $j$-cell in $S'$ and let $x\in S\setminus\{c',c\}$
  such that $\pi(x)=z$.  Then, $g(z)=x-(x:c)(c':c)c'$ if $j=i+1$ and
  $g(z)=x$ otherwise.
\item $\phi(c)=(c':c)c'$ and $\phi(x)=0$ for $x\in S\setminus\{c\}$.
\end{itemize}
To check that $(f,g,\phi)$ is a chain contraction of
$(C_*(S),\partial)$ to $(C_*(S'),\partial')$ is left to the reader.
\qed

\begin{definition}[Cells preserved-condition]
  Let $G$ be an $n$-Gmap. Let $c$ be an $i$-cell, and $E$ a subset of
  the graded cell $S_G$ (the set of all the cells of $G$).  We say
  that an operation on $G$ that provides a new $n$-Gmap $G'$ is
  $(E,c)-${\em preserved} if each $j$-cell $e\in E$ is after the
  operation a $j$-cell composed of darts $e\setminus c$.
\end{definition}

Note that in general, the contraction or the removal of a cell may
induce removal of other cells. 
For example, it is possible to build a sphere made of two vertices,
one codegree two edge and one face. Contracting the edge would
suppress all the darts and so the vertices and the face.  The cell
preserved condition ensures that when removing a degree two cell or
contracting a codegree two cell, other cells are preserved.

Now, the following proposition describes the condition to ensure that
removal and contraction preserve homology. In~\cite{DGP2012}, we focus
on the removal part of the following proposition. Here we generalize
this to removal and contraction.

\begin{proposition}\label{prop-degreetwo-preserve-homology}
  Let $c$ be a removable (resp. contractible) degree (resp. codegree)
  two $i$-cell in an \gmap{} $G$. Let $a$ and $b$ the two incident
  $(i+1)-$cells (resp. $(i-1)-$cells) of $c$, $E=S_G\setminus\{a,b,c\}$,
  and $G'$ the \gmap{} result of the operation.  If the removal
  (resp. contraction) of $c$ is $(E,c)-$preserved and if $a$ and $b$
  are merged into the $(i+1)-$cell (resp. $(i-1)-$cell) $a \cup b
  \setminus c$ in $G'$, then the homology groups of $G$
  and $G'$ are isomorphic.
\end{proposition}

Let us note $E_c$ the set of cells incident to $c$.  By definition of
removal and contraction operations, we know that the cells not in
$E_c$ are not modified by the operations. Thus we are sure that the
operation is $(S_G\setminus (E_c \cup \{a,b,c\}),c)-$preserved.  Thus
 the operation is $(E,c)-$preserved is equivalent to
saying that the operation is $(E_c\setminus\{a,b,c\},c)-$preserved:
the condition needs only to be verifyied for the cells incident to
$c$.

\proof Let us focus on the contraction part: 

It is immediate that $S_G' = S_G\setminus\{a,c\}$. 

As the operation is $(E,c)-$preserved, with $E=S_G\setminus\{a,b,c\}$,
there is bijection between the cells before and after the operation
$\pi : S_G\setminus\{a,b,c\} \rightarrow S_G'$ such that
$\pi(e)=e\setminus c$. We extend this bijection by defining
$\pi(b)=(a\cup b)\setminus c$.

Observe that $\partial_{G}(c)=(c:a)a+(c:b)b$ and for $e,x\in
S_{G}\setminus\{a,b,c\}$, we have that:
$$\begin{array}{l}
(e\setminus c:x\setminus c)=(e:x)\\
(e\setminus c:a\cup b\setminus c)=(e:b)-(e:a)(c:a)(c:b)\\
(a\cup b\setminus c:x\setminus c)=(e:b)
\end{array}$$ 
We have to prove that the boundary conditions in 
\refprop{prop-general-operation} are satisfied. Let $e$ be a $j$-cell in $S_G\setminus\{c,a\}$. We have to prove that:
$$\begin{array}{l}\mbox{$\partial_{G'}\pi(e)=\pi\left(\partial_G(e)-(e:c)c\right)$ if $j=i+1$;}\\
\mbox{$\partial_{G'}\pi(e)=\pi\left(\partial_G(e)-(e:a)(c:a)\partial_G(c)\right)$ if $j=i$,}\\
\mbox{$\partial_{G'}\pi(e)=\pi\partial_G(e)$ otherwise. }\end{array}$$

\begin{itemize}
\item  If  $j=i+1$ then
$\pi\left(\partial_G(e)-(e:c)c\right)=\sum_{x\in S_{G}\setminus \{a,b,c\}}(e:x)\pi(x)$\\
$=\sum_{x\in S_{G}\setminus \{a,b,c\}}(e:x)(x\setminus c)=\partial_{G'}(e\setminus c)=\partial_{G'}\pi(e)$.

\item If  $j=i$ then
$\pi\left(\partial_G(e)-(e:a)(c:a)\partial_G(c)\right)$\\
$=((e:b)-(e:a)(c:a)(c:b))\pi(b)+\sum_{x\in S_G\setminus\{a,b,c\}} (e:x)\pi(x)$\\
$=((e:b)-(e:a)(c:a)(c:b))(a\cup b\setminus c)+\sum_{x\in S_G\setminus\{a,b,c\}} (e:x)(x\setminus c)=\partial_{G'}(e\setminus c)=\partial_{G'}\pi(e)$.

\item If $j\neq i,i+1$ and $e=b$, then
$\pi\partial_G(b)=\sum_{x\in S_G\setminus\{a,b,c\}}(b:x)(x\setminus c)=\partial_{G'}(a\cup b\setminus c)=\partial_{G'}\pi(e)$.

\item  If  $j\neq i,i+1$  and $e\neq b$ then
$\pi\partial_G(e)=\sum_{x\in S_{G}\setminus\{a,b,c\}}(e:x)(x\setminus c)$\\
$=\partial_{G'}(e\setminus c)=\partial_{G'}\pi(e)$.
\end{itemize}

The prove is similar for removal operation.
\qed

\subsection{Dangling and Codangling cells}

Dangling and codangling cells are special cases as they do not satisfy the
degree/codegree two property. However they can also be simplified,
under some conditions, without modifying the homology of the $n$-Gmap.

Let $(C_*(S),\partial)$ be a chain complex.  Let $c$ be an $i$-cell
and $c'$ an $(i+1)$-cell, both in $S$, such that $|(c':c)|=1$,
$(x:c)=0$ for any $x\in S_{i+1}$, $x\neq c'$.  The operation under
which we remove $c$ and $c'$ from $S$ to get $S'=S\setminus \{c,c'\}$
is called {\em elementary collapse}.  By
\refprop{prop-general-operation}, the chain complexes $(C_*(S),\partial)$ and
$(C_*(S'),\partial|_S')$ have isomorphic homology groups, being $\pi:
S\setminus \{c,c'\}\to S'$ in this case, the identity.
Therefore an elementary collapse preserves homology. 

A subset $A$ of
$S$ is {\em collapsible} if all the elements of $A$ can be removed
from $S$ in a sequence of elementary collapses. That is, if we can
order the cells of $A$ as a sequence
$A=\{a_1,b_1,a_2,b_2,\dots,a_m,b_m\}$ such that
$S_i=S\setminus\{a_1,b_1,\dots,a_i,b_i\}$ is an elementary collapse of
$S_{i-1}= S_i\cup\{a_i,b_i\}$, for $1\leq i\leq m$.

Let $G$ be an $n$-Gmap and $c$  an $i$-cell in $G$. The {\em closure} of
$c$, denoted $\overline{c}$, is the set made of $c$ plus all the
$j$-cells, $0\leq j<i$ that are incident to $c$.  The closure of a set
$C$ of cells, denoted $\overline{C}$, is the union of the closures of
all the cells of $C$. Similarly, the {\em coclosure} of $c$, denoted
$\underline{c}$, is the set made of $c$ plus all the $j$-cells, $i<j
\leq n$ that are incident to $c$.  The coclosure of a set $C$ of
cells, denoted $\underline{C}$, is the union of the coclosures of all
the cells of $C$.

\begin{definition}[Dangling and codangling cells]\label{def-cellule-pendante} Let $c$
  be an $i$-cell in an $n$-Gmap. 
  \begin{itemize}
  \item Let $C$ be the set of $(i-1)-$cells incident to $c$, and
    $B=\{c'\in C | degree(c')>1\}$.  $c$ is \emph{dangling} if
    $degree(c)=1$ and $\{c\}\cup \overline{C}\setminus \overline{B} $
    is collapsible.
  \item Let $E$ be the set of $(i+1)-$cells incident to $c$, and
    $F=\{c'\in E | codegree(c')>1\}$.  $c$ is \emph{codangling} if
    $codegree(c)=1$ and $\{c\}\cup \underline{E}\setminus
    \underline{F} $ is collapsible.
  \end{itemize}
\end{definition}


In \cite{DGP2012} we stated that the removal of a removable dangling
cell preserves homology. A similar result holds for codangling cell.

\begin{proposition}\label{prop-dangling-preserve-homology}
  Let $c$ be an $i$-cell in an \gmap{} $G$.  
  \begin{itemize}
  \item If $c$ is removable and dangling cell, and the removal of $c$
    is $(\bar{B},c)-$preserved, then its removal preserves the
    homology of $G$.
  \item If $c$ is contractible and codangling cell, and the
    contraction of $c$ is $(\underline{F},c)$\-preserved, then its
    contraction preserves the homology of  $G$.
  \end{itemize}  
\end{proposition}

\proof   Let us prove the result for contractible codangling cells. Let $G'$ be
  the $n$-Gmap obtained after contracting $c$.  When we remove the
  $i$-cell $c$, then all the codegree one $(i+1)$-cells $e$ incident
  to $c$ are also removed from $G$ since all the darts of $e$ are
  darts of $c$. By the same reason, all the cells of
  $\underline{E}\setminus \underline{F} $ are removed from $G$ when we
  remove $c$ by the contraction operation.  No more cells are removed
  since the contraction of $c$ is $(\underline{F},c)-$preserved.
  Therefore, $S_{G'}=S_{G}\setminus (\underline{E}\setminus
  \underline{F})$.  Since $\underline{E}\setminus \underline{F} $ is
  collapsible, then $G$ and $G'$ have isomorphic homology groups.  The
  proof is similar for the removable dangling case.  \qed
 \qed

\section{Simplification Algorithm}
\label{sec:simplification-algorithm}

Now we can use the removal and contraction operations in order to
simplify a given \gmap{} \emph{G} while preserving its homology.  As
the number of cells in the simplified \gmap{} will be much smaller
than the number of cells in the initial one, we will speed-up the
homology computation by using the reduced \gmap{} instead of the
original one. Our simplification algorithm will start to remove cells,
then to contract cells.

The removal of $i$-cells which are either degree two or dangling cells
is presented in \refalgo{algo-removal-dimi}. As we have seen in the
previous section, these cells can be removed without modyfing the
homology of the \gmap{}.
%
\begin{algorithm2e}
  \KwIn{An \gmap{} \emph{G}.}
  \KwResult{Remove $i$-cells of \emph{G} while preserving the same homology.}
  \BlankLine

  \ForEach{$i$-cell $c$ of \emph{G}}
  {
    \If{$c$ is removable \textbf{and} the degree of $c$ is 2 
      \textbf{and} other cells are preserved}
    {Remove $c$\;}
    \lElse{\If{$c$ is removable \textbf{and} $c$ is a dangling cell}
      {Push($P,c$)\;
        \Repeat{empty($P$)}
        {$c \leftarrow$ pop($P$)\;
          \If{Other cells are preserved}
          {Push in $P$ all the removable dangling $i$-cells adjacent to $c$\;
            Remove $c$\;}
        }
      }
    }
  }
  
  \caption{Remove $i$-cells.}
  \label{algo-removal-dimi}
\end{algorithm2e}

In this algorithm, we iterate through all the $i$-cells of
\emph{G}. If the current cell $c$ is a removable degree two cell such
that the other cells are preserved, we remove it by using the removal
operation and we pass to the next $i$-cell. Otherwise, if $c$ is
removable, dangling and other cells are preserved, we also can remove
it. However we push in the stack $P$ all the dangling $i$-cells
adjacent to $c$. Indeed, these cells need to be reconsidered as they
are become dangling due to the removal of $c$.

%
\def\largFig{.22\textwidth}%
\begin{figure}
  \begin{center}
    \subfigure[\label{fig-simplif-dangling-a}]
    {\includegraphics[width=\largFig]{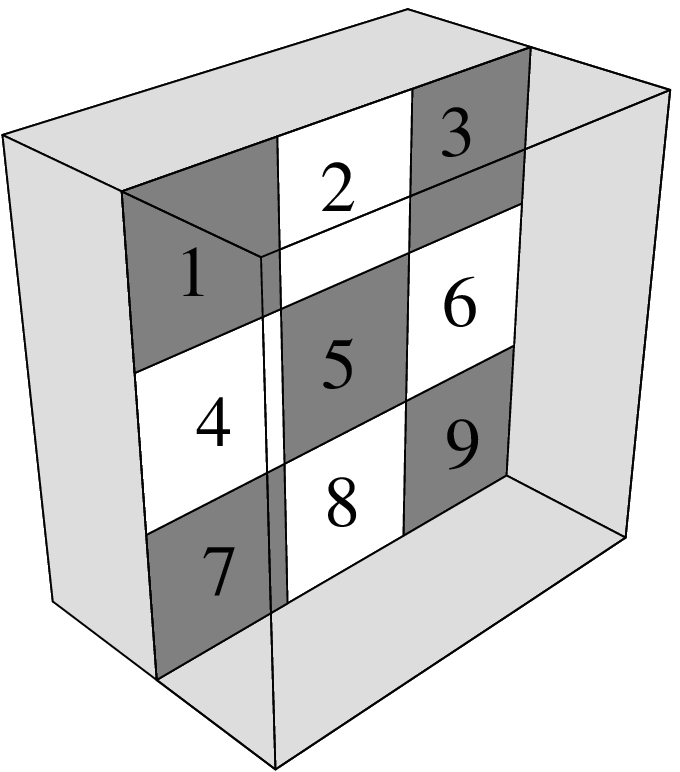}}\quad
    \subfigure[\label{fig-simplif-dangling-b}]
    {\includegraphics[width=\largFig]{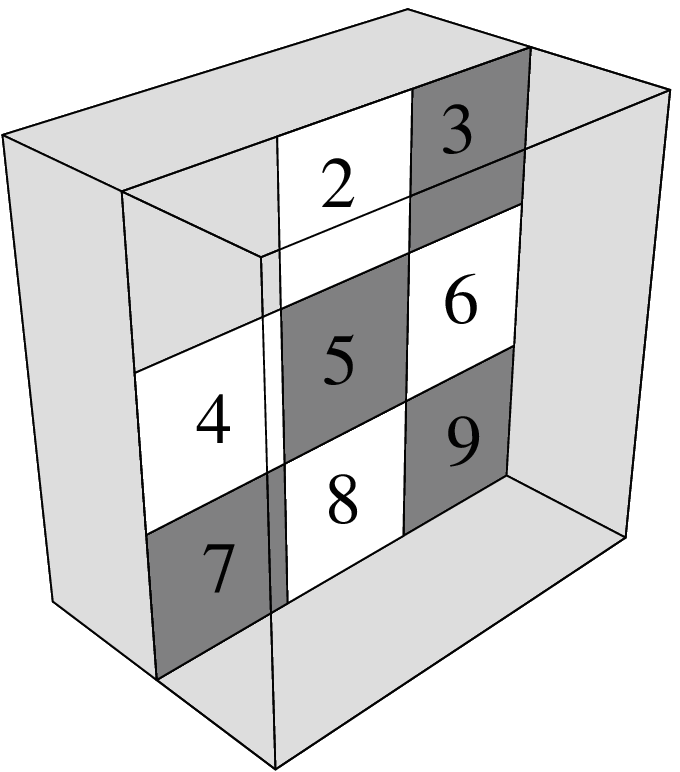}}\quad
    \subfigure[\label{fig-simplif-dangling-c}]
    {\includegraphics[width=\largFig]{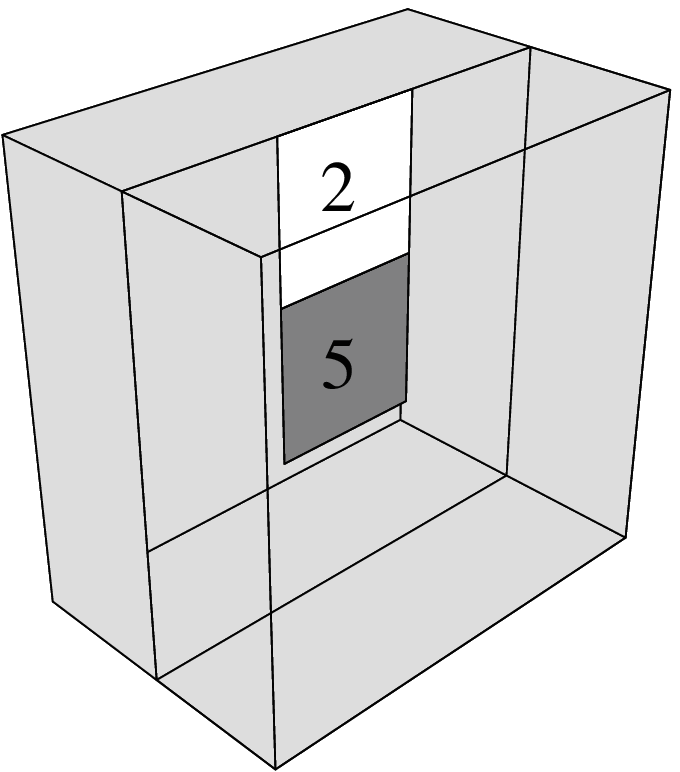}}\quad
    \subfigure[\label{fig-simplif-dangling-d}]
    {\includegraphics[width=\largFig]{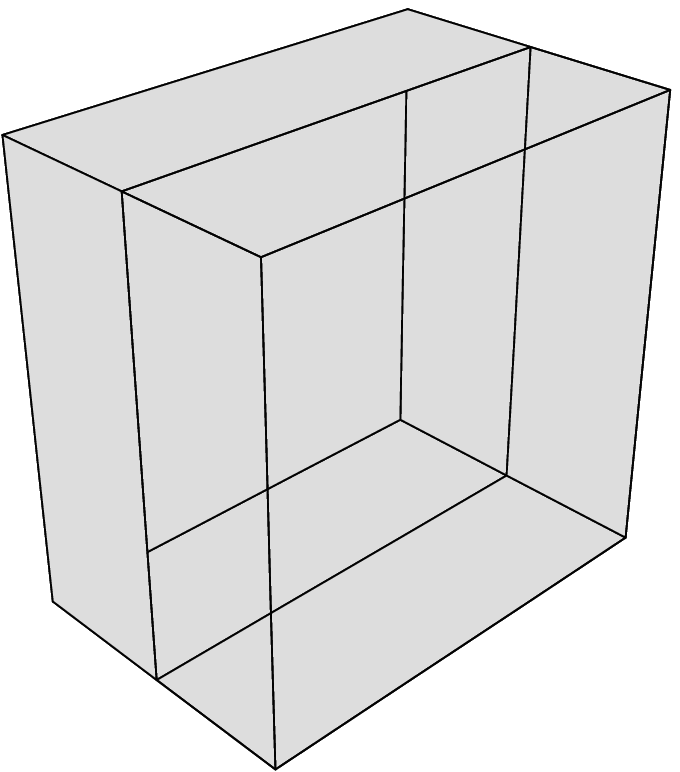}}
    \caption{Example to illustrate the removal of dangling cells.
      \subref{fig-simplif-dangling-a}~An initial configuration made of
      two volumes that share nine square faces numbered from 1 to 9.
      \subref{fig-simplif-dangling-b}~After the removal of face 1
      which was a removal degree two face.
      \subref{fig-simplif-dangling-c}~After the removal of faces 4, 6,
      8, 9, 6, 3 which were all removal dangling faces (when they are
      considered successively in this order). Here face 2 is not
      dangling while face 5 is. \subref{fig-simplif-dangling-c}~After
      the removal of face 5, face 2 becomes dangling and can be
      removed.}
    \label{fig-simplif-dangling}
  \end{center}
\end{figure} 
%
This case is illustrated in \reffig{fig-simplif-dangling} where we
start with a configuration made of two volumes that share nine square
faces numbered from 1 to 9 in \reffig{fig-simplif-dangling-a}.  As
faces as considered in any order, let us suppose we start to process
face number 1. This face is removable, has a degree equal to 2 and
cells are preserved, thus it is removed and we obtain the
configuration shown in \reffig{fig-simplif-dangling-b} where the two
cubes were merged, and all the faces numbered from 2 to 9 have now
degree 1. Then, let us suppose we consider faces numbered 4, 7, 8, 9,
6 and 3 in this order. Each face is removable and dangling, and thus
can be removed without modifying the homology of the 2-Gmap. We obtain
the configuration shown in \reffig{fig-simplif-dangling-c}. In this
configuration, if we consider face 2, this face is not dangling as 
the face plus the degree one edges in its boundary is not
collapsible. Then we consider face 5 which is dangling, and after its
removal, face 2 is now become dangling and thus be reconsidered a second
time.

We present in \refalgo{algo-contract-dimi} the similar algorithm for
contraction operation. 
%
\begin{algorithm2e}
  \KwIn{An \gmap{} \emph{G}.}
  \KwResult{Contract  $i$-cells of \emph{G} while preserving the same homology.}
  \BlankLine

  \ForEach{$i$-cell $c$ of \emph{G}}
  {
    \If{$c$ is contractible \textbf{and} the codegree of $c$ is 2
      \textbf{and} other cells are preserved}
    {Contract $c$\;}
    \lElse{\If{$c$ is contractible \textbf{and} $c$ is a codangling cell}
      {Push($P,c$)\;
        \Repeat{empty($P$)}
        {$c \leftarrow$ pop($P$)\;
          \If{Other cells are preserved}
          {Push in $P$ all the contractible codangling $i$-cells adjacent to $c$\;
          Contract $c$\;}
      }
      }
    }
  }
  
  \caption{Contract $i$-cells.}
  \label{algo-contract-dimi}
\end{algorithm2e}

Now we can use these two algorithms to simplify a given \gmap{}
\emph{G}: this global algorithm is given in
\refalgo{algo-simplification}. The principle of this global algorithm
is to start to remove $i$-cells, $i$ starting from $n-1$ (the
dimension of \emph{G} minus 1) and going downto 0. We consider the
cells in decreasing order for removal operation, as the removal of an
$i$-cell will decrease the degree of its incident $(i-1)$-cells. Thus
these cells could be non removable before the removal of the $i-$cell
and become removable after (as in the example in
\reffig{fig-removable-example}).  After all the removal operations, we
can continue the simplification by using the contraction
operation. Now we consider $i$-cell contractions starting from $i=1$
and going to $i=n$. Indeed, contracting an $i$-cell $c$ will decrease
the codegree of its $(i+1)$-cells thus these cells could become
contractible after the contraction of $c$.
%
\begin{algorithm2e}
  \KwIn{An \gmap{} \emph{G}.}
  \KwResult{Simplify \emph{G} while preserving the same homology.}
  \BlankLine

  \For{$i \leftarrow n-1$ to 0}
  {
    Remove $i$-cells\;
  }
  \For{$i \leftarrow 1$ to $n$}
  {
    Contract $i$-cells\;
  }
  
  \caption{Simplification of a given \gmap{}.}
  \label{algo-simplification}
\end{algorithm2e}

Notice that there are no particular arguments to start to remove then to
contract cells, and we can inverse these two steps without problem and
obtain also a simplified \gmap{} having the same homology than
\emph{G}.

\emph{Complexity:} The two algorithms \refalgo{algo-removal-dimi} and
\refalgo{algo-contract-dimi} have a complexity linear in number of
darts of \emph{G}. Indeed, considering all the $i$-cells can be done
linearly in number of darts by using a Boolean mark to mark darts
already considered. The two tests of being removable/contractible, and
the degree/codegree computation have a complexity linear in number of
darts of the considered cell, and in number of darts in its incidence
cells. Removal and contraction operations are linear in number of
darts of the cell. Lastly, we are sure that reconsidered cells are
retested only once as they become dangling, they are now removed the
second time they are treated.

The complexity of the global simplification method
\refalgo{algo-simplification} is thus linear in the number of darts of
\emph{G} times the dimension of the space (which is a constant
number). Moreover, notice that each successive step of remove
$i$-cells or contract $i$-cells is quicker than the previous one as
the number of darts decreases after each new simplification step.

\section{Experiments}
\label{sec:expe}

We have implemented our simplification algorithm and the computation
of the homology generators of a Gmap in \texttt{Moka} \cite{VidDam03},
a 3D topological modeler based on a kernel made of 3-Gmaps. In the
current version of our code, the simplification of removable degree
two cells, dangling cells and edge contraction of degree two
contractible cells have been implemented.  We are working on the code
to implement the face contraction, and the case of codangling cells
but this is not finished yet. However, even with this limited version,
we already have interesting results illustrating the interest of
simplifying contractible cells in addition to removable cells.

To compute homology generators, we compute incidence matrices (which
describe the boundary of the cells) using the signed incidence number
between all the cells of the \gmap{}. Then we reduce incidence
matrices into their Smith-Agoston normal form to compute homology
generators~\cite{Ago76}.  In this Agoston reduced normal form, for a
given dimension $d$, the basis of the boundaries $B_p$ is a subset of
the basis of cycles $Z_p$, thus the quotient group $H_p={Z_p}/{B_p}$
can directly be obtained by simply removing from $Z_p$ the boundaries
of infinite order. Note that by using the definition of removal and
contraction operations, we are able to project the generators of the
simplified object on the initial one.

In \cite{DGP2012}, we have made some experiments where we compared the
results obtained by our method which computes the homology of the
simplified objects using removal operations only, with other two 
methods \texttt{Chomp} \cite{Chomp} and \texttt{RedHom}
\cite{RedHom}. The results of these experiments show that our method
was generally quicker than both other methods. As in this paper we
improve the previous method given in \cite{DGP2012}, we only make some
experiments to compare the new approach with this previous one.

Thus we present here the results of two different experiments that
illustrate the generality of our method. In a first experiment, we
compute the 2D homology generators of 320 2D triangular meshes
described by 2-Gmaps. These meshes are taken from a 3D database
available in the Shape Retrieval Contest web page \cite{Shrec}.  In a
second experiment, we have computed the 3D homology generators of 300
3D set of voxels. Each set of voxels is randomly generated within an
image of size $64^3$. In the first case, we compute 2D simplicial
homology while in the second case we compute 3D cubical homology. In
both cases, we use the same code which shows the interest of using a
generic framework allowing to represent any type of cells.

\subsection{2D Triangular Models}

We present in \reffig{fig-exemple-mesh} some 3D meshes extracted from
the \texttt{Shrec} database, and in \reftab{tab-stats-expe2d} some
characteristics of the 320 objects used in this experiment. Note that
all these objects are orientable, thus there is no torsion in the
homology groups. Moreover, as each face is a triangle, there is no
0-free nor 1-free darts, but there are sometimes some 2-free darts for
meshes with boundary.
%
\def\LargFig{.31\textwidth}%
\begin{figure}
 \begin{center}
    \subfigure[]{\includegraphics[width=\LargFig]{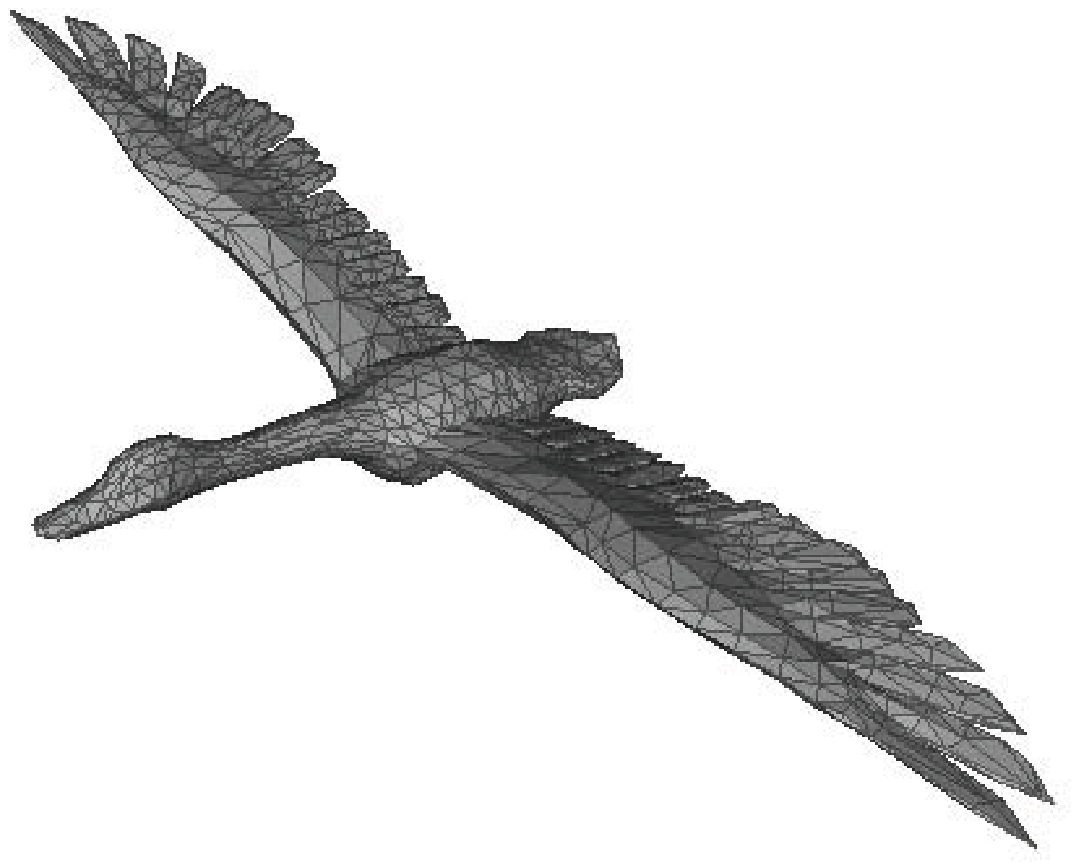}}\quad
    \subfigure[]{\includegraphics[width=\LargFig]{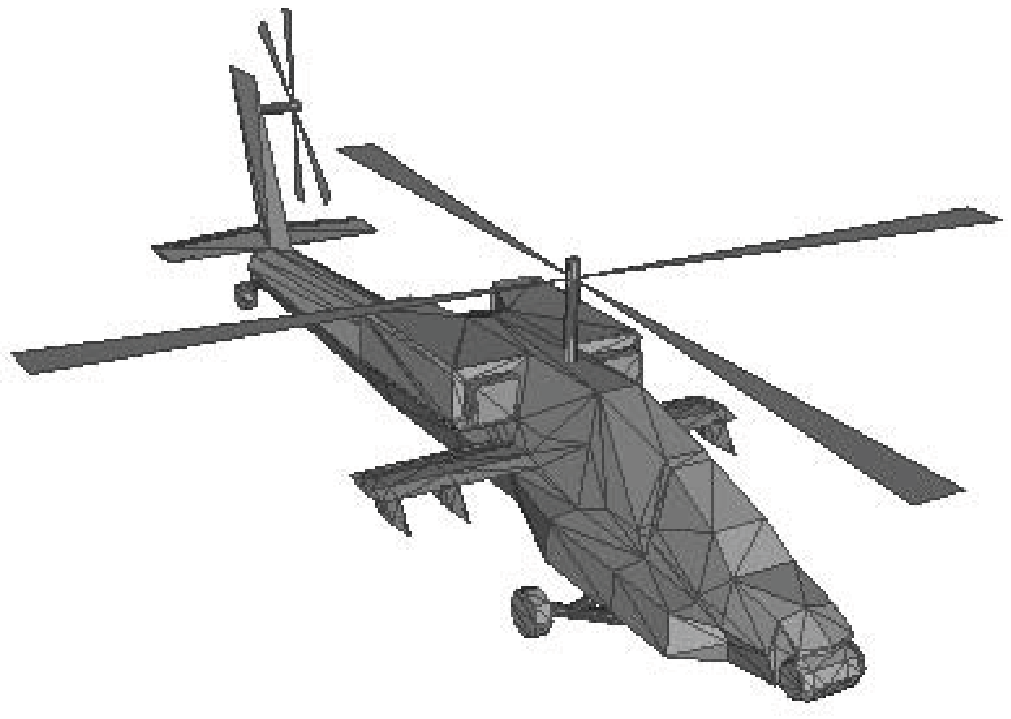}}\quad
    \subfigure[]{\includegraphics[width=\LargFig]{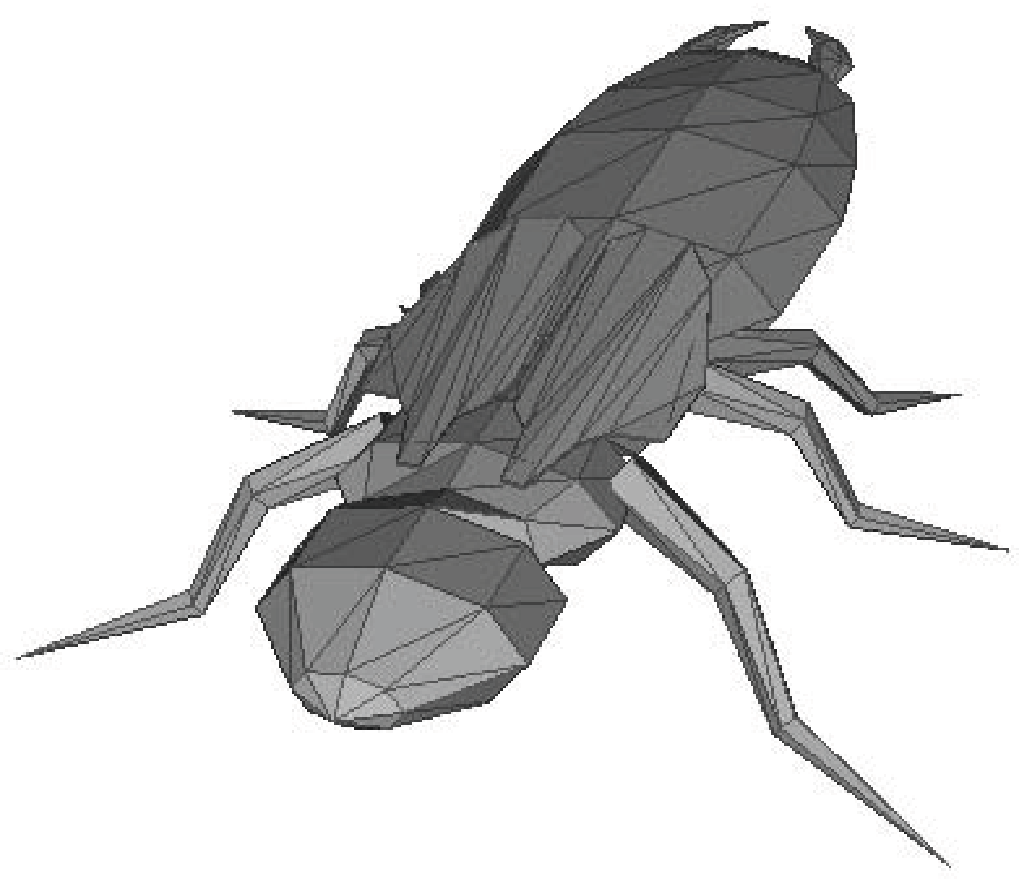}}
 \end{center}
 \caption{Examples of some meshes from the \texttt{Shrec} database.}
 \label{fig-exemple-mesh}
\end{figure}

\begin{table}
\begin{center}
\begin{tabular}{||r||r|rrr|rrr||}
\hline
     & \multicolumn{1}{c|}{\# darts} & 
      \multicolumn{1}{c}{$|S_G^0|$} &
      \multicolumn{1}{c}{$|S_G^1|$} & 
      \multicolumn{1}{c|}{$|S_G^2|$} & 
      \multicolumn{1}{c}{$B_0$} & 
      \multicolumn{1}{c}{$B_1$} & 
      \multicolumn{1}{c||}{$B_2$} \\
\hline
min  & 696    & 72    & 184    & 116   & 1    & 0    & 0 \\
max  & 540966 & 47000 & 137125 & 90161 & 3221 & 1416 & 1321 \\
mean & 69660  & 6631  & 18045  & 11610 & 190  & 35   & 41 \\
std  & 84391  & 7577  & 21472  & 14065 & 446  & 123  & 135 \\
\hline
\end{tabular}
\end{center}
\caption{Characteristics of the 320 objects used in our 2D experimentation. The
  first 4 columns give the number of darts, vertices, edges and faces.
  (the number of $i$-cells is the cardinal of the
  set $S_G^i$). The 3 last columns give the Betti numbers $B_0$, $B_1$
  and $B_2$.  For each characteristic, we give the minimum and maximum
  value, the mean and the standard deviation.}
\label{tab-stats-expe2d}
\end{table}

For each object, we have first simplified the $2$-Gmap by using
removal operations only, and we have computed the time required to
simplify this Gmap, the characteristics of the simplified Gmap and the
time required to compute its homological generators.  Then, starting
from the same initial object, we have simplified the $2$-Gmap by using
removal and contraction operations, and have computed the same
values. This allows us to show the interest of using contraction
simplifications in addition to removal ones.

%
\begin{table}
\begin{center}
\begin{tabular}{||r||r|rrr|rr||}
\hline
\multicolumn{7}{||c||}{Removal only}\\
\hline
  & & & & & \multicolumn{2}{c||}{Time} \\
  & \multicolumn{1}{c|}{\# darts} &
  \multicolumn{1}{c}{$|S_G^0|$} &
  \multicolumn{1}{c}{$|S_G^1|$} & 
  \multicolumn{1}{c|}{$|S_G^2|$} & 
  \multicolumn{1}{c}{Simplif.} & 
  \multicolumn{1}{c||}{Homology} \\
\hline
min  & 2      &    1&   1&   1 & 0    & 0       \\
max  & 17046  & 5631&5726&3221 & 0,27 & 6435,41 \\
mean & 722    &  282& 276& 190 & 0,04 & 47,40   \\
std  & 1621   &  620& 619& 446 & 0,05 & 391,83  \\
\hline
\end{tabular}
\begin{tabular}{||r||r|rrr|rr||}
\hline
\multicolumn{7}{||c||}{Removal and Contraction}\\
\hline
  & & & & & \multicolumn{2}{c||}{Time} \\
  & \multicolumn{1}{c|}{\# darts} &
  \multicolumn{1}{c}{$|S_G^0|$} &
  \multicolumn{1}{c}{$|S_G^1|$} & 
  \multicolumn{1}{c|}{$|S_G^2|$} & 
  \multicolumn{1}{c}{Simplif.} & 
  \multicolumn{1}{c||}{Homology} \\
\hline
min  & 2     &  1   &   1  &   1  &  0    & 0       \\
max  & 15056 & 4700 & 4795& 3221 &  0,28 & 3644,35 \\
mean & 653   &  257 & 251 & 190  &  0,04 & 33,87   \\
std  & 1463  &  569 & 559 & 446  &  0,05 & 244,59  \\
\hline
\end{tabular}
\end{center}
\caption{Results of our 2D experiments. We give the number of cells
  (columns $|S_G^i|$), the simplification time (columns
  \emph{Time$\backslash$Simplif.})  and the computation time for
  homology generators (columns \emph{Time$\backslash$Homology}) for the
  objects simplified by using removal operations only, then for objects
  simplified by using removal and contraction operations.  Times are
  given in seconds, 0s means less than $10^{-6}$s.}
\label{tab-res-expe2d}
\end{table}
%

The results are shown in \reftab{tab-res-expe2d}. We can see in these
results that in average, there are 25 edges that are contracted, which
represents about 10\% of the total number of edges. Note that as we do
not contract faces, nor codangling cells, the number of faces is not
modified by the contraction simplification. We can notice that the
time spend by the simplification process is near equal between the two
versions. This can be explained by the small number of cells in the
2-Gmap after the removal simplification. However, there is a non
negligible gain for the time spend to compute the homology generators:
in average about 14 seconds which is about 29\%.  Note that in
\reftab{tab-res-expe2d} we only present the Betti numbers, but in
practice we compute homology generators which give more information.

These results show the interest of using contraction simplification to
speed up the computation of homology generators. Moreover, this
interest is more important for bigger objects. For example, the
maximum time spend for the computation of homology generators is 6435
seconds if we use the removal operations only, while it is 3644
seconds if we use both removal and contraction operations.

\subsection{3D Set of Voxels}

In this second experiment, we generate randomly 300 3D set of voxels
within an image of size $64^3$. Then as in the previous experiment, we
compare the results obtained by the simplification method using only
removal operations with the results obtained by the simplification
method using removal and contraction operations.  As in the previous
experiment, all the objects generated here are orientable and thus
there is no torsion coefficient.  Moreover, as the voxels are embedded
in 3D Euclidean space, the homology group $H_3$ is always trivial and
thus the corresponding Betti number $B_3=0$.

We give in \reftab{tab-stats-expe3d} the characteristics of the 3D
generated objects.  As we use the same method to randomly create all
the objects, we can see that they have all similar number of darts and
cells. However, as the position of voxels is randomly chosen, we have
different Betti numbers.
\begin{table}
\begin{center}
\begin{tabular}{||r||r|rrrr|rrr||}
\hline
     & \multicolumn{1}{c|}{\# darts} & 
      \multicolumn{1}{c}{$|S_G^0|$} &
      \multicolumn{1}{c}{$|S_G^1|$} & 
      \multicolumn{1}{c}{$|S_G^2|$} & 
      \multicolumn{1}{c|}{$|S_G^3|$} & 
      \multicolumn{1}{c}{$B_0$} & 
      \multicolumn{1}{c}{$B_1$} & 
      \multicolumn{1}{c||}{$B_2$} \\
\hline
min  & 12227904&271780&797654&780869&254748&  1& 0&0 \\
max  & 12582768&274628&811198&798716&262141&150&44&7 \\
mean & 12422916&273554&805519&790809&258811& 49&18&2 \\
std  & 81516   &   654&  3162&  4190&  1698& 34&10&2 \\
\hline
\end{tabular}
\end{center}
\caption{Characteristics of the 300 objects used in our 3D
  experimentation. The first 5 columns give the number of darts,
  vertices, edges, faces and volumes (the number of $i$-cells is the
  cardinal of the set $S_G^i$). The 3 last columns give the Betti
  numbers $B_0$, $B_1$ and $B_2$ ($B_3$ is always 0).  For each
  characteristic, we give the minimum and maximum value, the mean and
  the standard deviation.}
\label{tab-stats-expe3d}
\end{table}

The results of our method that computes the homology generators for
these 3D cubical objects are given in \reftab{tab-res-expe3d}.
Firstly we must notice that the number of cells is significantly
decreased by both simplification methods. This can be characterize by
the number of darts which in average starts from $12,422,916$ and
decreases to $1,435$ for removal only and to $1,273$ for removal and
contraction.  Secondly,  the number of contracted edges is in
average 40, which represents about 18\% of the total number of edges.
This explains the gain for the computation time of homology generators
for the method with removal and contraction operations which is in
average $0.41$ seconds, about 10\% of the time of the removal only
simplification method.
%
\begin{table}
\begin{center}
\begin{tabular}{||r||r|rrrr|rr||}
\hline
\multicolumn{8}{||c||}{Removal only}\\
\hline
  & & & & & & \multicolumn{2}{c||}{Time} \\
  & \multicolumn{1}{c|}{\# darts} &
  \multicolumn{1}{c}{$|S_G^0|$} &
  \multicolumn{1}{c}{$|S_G^1|$} & 
  \multicolumn{1}{c}{$|S_G^2|$} & 
  \multicolumn{1}{c|}{$|S_G^3|$} & 
  \multicolumn{1}{c}{Simplif.} & 
  \multicolumn{1}{c||}{Homology} \\
\hline
min  &   28&  5&  5&  3&  1 &3,30 & 0,00\\
max  & 4152&497&563&270&150 &3,93 & 4,86\\
mean & 1435&198&215& 99& 49 &3,63 & 0,50\\
std  &  810&109&115& 56& 34 &0,09 & 0,73\\
\hline
\end{tabular}
\begin{tabular}{||r||r|rrrr|rr||}
\hline
\multicolumn{8}{||c||}{Removal and Contraction}\\
\hline
  & & & & & & \multicolumn{2}{c||}{Time} \\
  & \multicolumn{1}{c|}{\# darts} &
  \multicolumn{1}{c}{$|S_G^0|$} &
  \multicolumn{1}{c}{$|S_G^1|$} & 
  \multicolumn{1}{c}{$|S_G^2|$} & 
  \multicolumn{1}{c|}{$|S_G^3|$} & 
  \multicolumn{1}{c}{Simplif.} & 
  \multicolumn{1}{c||}{Homology} \\
\hline
min  &   16&  2&  2&  3&1   & 3,27 & 0,00\\ 
max  & 3868&434&494&270&150 & 3,52 & 3,57\\
mean & 1273&158&175& 99&49  & 3,36 & 0,35\\
std  &  751& 95& 99& 56&34  & 0,04 & 0,53\\
\hline
\end{tabular}
\end{center}
\caption{Results of our 3D experiments. We give the number of cells
  (columns \emph{\# $i$-cells}), the simplification time (columns
  \emph{Time$\backslash$Simplif.})  and the computation time for
  homology generators (columns \emph{Time$\backslash$Homology}) for the
  objects simplified by using removal operations only, then for objects
  simplified by using removal and contraction operations.  Times are
  given in seconds, 0s means less than $10^{-6}$s.}
\label{tab-res-expe3d}
\end{table}
%

The gain is here less important than for our 2D experiments, 10\%
instead of 29\%. This can be explain by the fact that we did not use
yet the face and volume contractions.  In 2D, edge contraction have a
more relative impact as edges are used in the two incidence matrices,
while in 3D, edges are used in two incidence matrices among
three. For this reason, we think we could improve significantly our 3D
results by using face and volume contractions.

\section{Conclusion}
\label{sec:conclu}

In this paper, we have provided two new propositions giving the
conditions allowing to contract codegree two and codangling cells, and
we have proven that under these conditions, the homology of the
$n$-Gmap is preserved. By using similar result than in \cite {DGP2012}
for removal operations, we have proposed an algorithm which simplifies
a given $n$-Gmap by removing cells by decreasing dimension, then
contracting cells by increasing dimension. Thanks to our propositions,
we know that the homology of the Gmap is preserved during all the
simplification process. Thus we can compute the homology generators on
the simplified objects. The computation is faster as the number of
cells of the simplified objects is small.

To show the interest of doing more simplifications, we have make two
experiments to compare the results of the computation of homology
generators when we simplify the objects by using only removal
operations and by using removal and contraction operations. Even if
the method is not fully implemented (in the current version of our
code we only contract edges), the results show a non negligible gain
when the objects are more simplified. Moreover, there is almost no
overhead for the contraction step due to the fact that the object has
already a small number of cells after the removal step.

These experiments show also the interest of using a model that allows
to describe any type of cells: with the same software we are able to
compute simplicial and cubical homology generators in 2D and in 3D.

Our first perspective is to finish the implementation of contraction
operations for faces and volumes, and for codangling cells. We hope we can
 improve again our results as the objects will be more simplified.
We also want to make some experiments in higher dimensions with
orientable and non orientable objects.  Then, we can study if we can
propose other simplification operations that preserve homology.

\bibliographystyle{plain}
\bibliography{cviu-damiand-gonzalezdiaz-peltier}

\begin{thebibliography}{10}

\bibitem{Ago76}
M.~K. Agoston.
\newblock {\em Algebraic Topology, a first course}.
\newblock Pure and applied mathematics. Marcel Dekker Ed., 1976.

\bibitem{AlayranguesAl11}
S.~Alayrangues, G.~Damiand, P.~Lienhardt, and S.~Peltier.
\newblock A boundary operator for computing the homology of cellular
  structures.
\newblock {\em Discrete \& Computational Geometry}, under submission.

\bibitem{AlayranguesAl09}
S.~Alayrangues, S.~Peltier, G.~Damiand, and P.~Lienhardt.
\newblock Border operator for generalized maps.
\newblock In {\em Proc. of Discrete Geometry for Computer Imagery}, volume 5810
  of {\em LNCS}, pages 300--312, Montr{\'e}al, Canada, September 2009. Springer
  Berlin/Heidelberg.

\bibitem{basak10}
T.~Basak.
\newblock Combinatorial cell complexes and poincar{\'e} duality.
\newblock {\em Geometriae Dedicata}, 147:357--387, 2010.

\bibitem{Chomp}
Chomp.
\newblock http://chomp.rutgers.edu/.

\bibitem{DGP2012}
G.~Damiand, R.~Gonzalez-Diaz, and S.~Peltier.
\newblock Removal operations in nd generalized maps for efficient homology
  computation.
\newblock In {\em Proc. of International Workshop on Computational Topology in
  Image Context}, volume 7309 of {\em LNCS}, pages 20--29, Bertinoro, Italy,
  May 2012. Springer Berlin/Heidelberg.

\bibitem{DamLie03}
G.~Damiand and P.~Lienhardt.
\newblock Removal and contraction for n-dimensional generalized maps.
\newblock In {\em Proc. of Discrete Geometry for Computer Imagery}, volume 2886
  of {\em LNCS}, pages 408--419, Naples, Italy, November 2003. Springer
  Berlin/Heidelberg.

\bibitem{pawel}
P.~D\l{}otko, T.~Kaczynski, M.~Mrozek, and T.~Wanner.
\newblock Coreduction homology algorithm for regular cw-complexes.
\newblock {\em Discrete \& Computational Geometry}, 46:361--388, 2011.

\bibitem{DHSW03}
J.-G. Dumas, F.~Heckenbach, B.~D. Saunders, and V~Welker.
\newblock Computing simplicial homology based on efficient smith normal form
  algorithms.
\newblock In {\em Algebra, Geometry, and Software Systems}, pages 177--206,
  2003.

\bibitem{EdelsbrunnerLZ02}
Herbert E., David L., and Afra Z.
\newblock Topological persistence and simplification.
\newblock {\em Discrete {\&} Computational Geometry}, 28(4):511--533, 2002.

\bibitem{Hat02}
A.~Hatcher.
\newblock {\em Algebraic Topology}.
\newblock Cambridge University Press, 2002.
\newblock available on
  http://www.math.cornell.edu/$\sim$hatcher/AT/ATpage.html.

\bibitem{KMS98}
T.~Kaczyinski, M.~Mrozek, and M.~Slusarek.
\newblock Homology computation by reduction of chain complexes.
\newblock {\em Computers \& Mathematics with Applications}, 35(4):59 -- 70,
  1998.

\bibitem{KMM04}
T.~Kaczynski, K.~Mischaikow, and M.~Mrozek.
\newblock {\em Computational Homology}.
\newblock Springer, 2004.

\bibitem{Lienhardt91}
P.~Lienhardt.
\newblock Topological models for boundary representation: a comparison with
  n-dimensional generalized maps.
\newblock {\em Commputer Aided Design}, 23(1):59--82, 1991.

\bibitem{Lienhardt94}
P.~Lienhardt.
\newblock N-dimensional generalized combinatorial maps and cellular
  quasi-manifolds.
\newblock {\em Computational Geometry \& Applications}, 4(3):275--324, 1994.

\bibitem{mclane}
S.~MacLane.
\newblock {\em Homology}.
\newblock Classic in Mathematics. Springer, 1995.

\bibitem{May67}
J.~P. May.
\newblock {\em Simplicial objects in algebraic topology}.
\newblock Van Nostrand, Princeton, 1967.

\bibitem{morozov}
N.~Milosavljevi\'{c}, D.~Morozov, and P.~Skraba.
\newblock Zigzag persistent homology in matrix multiplication time.
\newblock In {\em Proc. of the 27th annual ACM symposium on Computational
  geometry}, SoCG '11, pages 216--225, New York, NY, USA, 2011. ACM.

\bibitem{Mun84}
J.~R. Munkres.
\newblock {\em {E}lements of algebraic topology}.
\newblock Perseus Books, 1984.

\bibitem{NSKPMT02}
M.~Niethammer, A.N. Stein, W.D. Kalies, P.~Pilarczyk, K.~Mischaikow, and
  A.~Tannenbaum.
\newblock Analysis of blood vessel topology by cubical homology.
\newblock In {\em IEEE Proceedings of the International Conference on Image
  Processing}, volume~2, pages 969--972, 2002.

\bibitem{PAFL06}
S.~Peltier, S.~Alayrangues, L.~Fuchs, and J.-O. Lachaud.
\newblock Computation of homology groups and generators.
\newblock {\em Computers \& Graphics}, 30:62--69, febuary 2006.

\bibitem{RedHom}
Redhom.
\newblock http://redhom.ii.uj.edu.pl/.

\bibitem{Shrec}
Shrec.
\newblock http://www.aimatshape.net/event/SHREC/.

\bibitem{Storjohann}
A.~Storjohann.
\newblock Near optimal algorithms for computing smith normal forms of integer
  matrices.
\newblock In {\em Proc. of the 1996 international symposium on Symbolic and
  algebraic computation}, ISSAC '96, pages 267--274, New York, NY, USA, 1996.
  ACM.

\bibitem{VidDam03}
F.~Vidil and G.~Damiand.
\newblock Moka.
\newblock http://moka-modeller.sourceforge.net/, 2003.

\end{thebibliography}
\appendix 
\label{proof-prop-4}

\section{Proof of \refprop{prop-general-operation}}

By assumption, let $(x:y)=0$ if $degree(x)-degree(y)\neq 1$. The following property will be used throughout the proof: 
$$\mbox{for any $(i+2)-$cell $x$, }\; \sum_{y\in S_{i+1}}(x:y)(y:c)=0$$ by the condition $\partial\partial(x)=0$.  

Now, let us check that $f$ is a chain map. Let $x\in S$, then:
\begin{itemize}
\item If $x=c$, then $\partial' f(c)=\partial' \pi(c-(c':c)\partial(c'))=\pi \partial(c-(c':c)\partial(c'))=\pi\partial(c)$\\
$=f\partial(c)$. 
\item If $x =c'$ then $f\partial(c`)=f\left((c':c)c+\sum_{y\neq c} (c':y)y\right)$\\
$=(c':c)\pi(c-(c':c)\partial(c'))+\sum_{y\neq c} (c':y)\pi(y)$\\
$=\pi((c':c)c-\partial(c'))+\sum_{y\neq c} (c':y)\pi(y)=0$.
\item Let $x$ be a $j$-cell in $S\setminus\{c,c'\}$.
\begin{itemize}
\item If $j=i+2$ then $\partial' f(x)=\partial'\pi(x)=\pi\left(\partial(x)-(x:c')c'\right)$\\
$=\sum_{y\in S\setminus\{c , c'\}}(x:y)\pi(y)=\sum_{y\in S\setminus\{c , c'\}}(x:y)y=f\partial(x)$.
\item If $j=1+1$, then $\partial' f(x)=\partial'\pi(x)=\pi\left(\partial(x)-(x:c)(c':c)\partial(c')\right)$\\
$=\pi\left(\partial(x)-(x:c)c+(x:c)c-(x:c)(c':c)\partial(c')\right)$\\
$=f(\partial(x)-(x:c)c)+(x:c)f(c)=f\partial(c)$.
\item
If $j\neq i+1,i+2$ then $f\partial(x)=\pi\partial(x)=\partial\pi(x)=\partial f(x)$.
\end{itemize}
\end{itemize}

Now, let us see that $g$ is a chain map. Let $z$ be a $j$-cell in $S'$, then there exists a $j$-cell $x\in S$ such that $\pi(x)=z$. Then:
$g\partial'(z)=g\partial'\pi(x)$.
\begin{itemize}
\item If $j=i+2$ then $g\partial'\pi(x)=g\pi\left(\partial(x)-(x:c')c'\right)$\\
$=\sum_{y\in S\setminus\{c , c'\}}(x:y)g\pi(y)=\sum_{y\in S\setminus\{c , c'\}}(x:y)(y-(y:c)(c':c)c')$\\
$=\sum_{y\in S\setminus\{c , c'\}}(x:y)y+(x:c')c'=\partial(x)=\partial g(z)$.

\item If $j=i+1$ then $g \partial' \pi(x)=g\pi (\partial(x)-(x:c)(c':c)\partial(c'))$\\
$=\partial(x)-(x:c)(c':c)\partial(c')=\partial(x-(x:c)(c':c)c') =\partial g(z)$.

\item If $j\neq i+1,i+2$ then $g\partial' (z)=g\partial'\pi(x)=g\pi\partial(x)$\\
 $=\sum_{y\in S\setminus\{c , c'\}}(x:y)g\pi(y)= \sum_{y\in S\setminus\{c , c'\}}(x:y)y-(y:c)(c':c)c'$\\
 $= \sum_{y\in S\setminus\{c , c'\}}(x:y)y=\partial(x)=\partial g(z)$.
\end{itemize}

Now, let us check that $fg=id_{C_*(S')}$. Let $z\in S'$ and $x\in S\setminus\{c,c'\}$ such that $\pi(x)=z$. Then
$fg(z)=fg(\pi(x))=f(x-(x:c)(c':c)c')=\pi(x)=z$.

Finally, let us see that $id_{C_*(S)}=gf+\phi\partial+\partial\phi$. Let $x$ be a $j$-cell in $S$:
\begin{itemize}
\item If $x=c$ then
$gf(c)=g\pi(c-(c':c)\partial(c'))=c-(c':c)\partial(c')$;
$\phi\partial(c)=0$
and $\partial\phi(c)=(c':c)\partial(c')$.
Then
$gf(c)+\phi\partial(c)+\partial\phi(c)=c$.
\item If $x=c'$ then
$gf(c')=0$; $\phi\partial(c')=(c':c)(c':c)c=c$ and  $\partial\phi(c')=0$. Then 
$gf(c')+\phi\partial(c')+\partial\phi(c')=c'$.
\item If $x\neq c,c'$ then 
$gf(x)=g\pi(x)=x-(x:c)(c':c)c'$; $\partial\phi(x)=0$ and
$\phi\partial(x)=(x:c)(c':c)c'$. Then 
$gf(x)+\phi\partial(x)+\partial\phi(x)=x$.
\end{itemize}

\section{Proof of \refprop{prop-degreetwo-preserve-homology} Removable Case}

 Let $c$ be a removable degree two $i$-cell in an \gmap{} $G$.
  Let $a$ and $b$ its two incident $(i+1)-$cells.
$\pi:S_{G}\setminus \{c,a\}\to S_{G'}$ is defined as: 
$\pi(e)=e\setminus c$ for any $j$-cell $e\in S_{G}\setminus \{c,a,b\}$;
   and $\pi(b)=a\cup b\setminus c$.

Observe that for $e,x\in S_{G}\setminus\{a,b,c\}$,    we have that :
$$\begin{array}{l}
(e\setminus c:x\setminus c)=(e:x)\\
(e\setminus c:a\cup b\setminus c)=(e:b)\\
(a\cup b\setminus c:x\setminus c)=(b:x)-(b:c)(a:c)(a:x)
\end{array}$$ 

We have to prove that the boundary conditions in 
\refprop{prop-general-operation} are satisfied. Let $e$ be a $j-$cell in $S_G\setminus\{a,c\}$. We have to prove that. 
$$\begin{array}{l}
\mbox{$\partial_{G'}\pi(e)=\pi\left(\partial_G(e)-(e:a)a\right)$ if $j=i+2$,}\\
\mbox{$\partial_{G'}\pi(e)=\pi\left(\partial_G(e)-(e:c)(a:c)\partial_G(a)\right)$ if $j=i+1$,}\\
\mbox{$\partial_{G'}(e)=\pi\partial_G(e)$ otherwise.}
\end{array}$$

\begin{itemize}
\item  If  $j=i+2$ then
$\pi\left(\partial_G(e)-(e:a)a\right)$\\
$=(e:b)\pi(b)+\sum_{x\in S_G\setminus\{a,b,c\}}(e:x)\pi(x)$\\
$=(e:b)(a\cup b\setminus c)+\sum_{x\in S_G\setminus\{a,b,c\}}(e:x)(x\setminus c)=\partial_{G'}(e\setminus c)=\partial_{G'}\pi(e)$. 

\item If $j=i+1$ and $e=b$ then
$\pi\left(\partial_G(b)-(b:c)(a:c)\partial_G(a)\right)$\\
$=\sum_{x\in S_G\setminus\{a,c,b\}}((b:x)-(b:c)(a:c)(a:x))\pi(x)$\\
$=\sum_{x\in S_G\setminus\{a,c,b\}}((b:x)-(b:c)(a:c)(a:x))(x\setminus c)
=\partial_{G'}(a\cup b\setminus c)$\\
$=\partial_{G'}\pi(b)$.

\item If  $j=i+1$ and $e\neq b$, since $e\neq a$ then $(e:c)=0$, then\\
$\pi\left(\partial_G(e)-(e:c)(a:c)\partial(a)\right)=\sum_{x\in S_G\setminus\{a,c,b\}}(e:x)\pi(x)$\\
$=\sum_{x\in S_G\setminus\{a,c,b\}}(e:x)(x\setminus c)=\partial_{G'}(e\setminus c)=\partial_{G'}\pi(e)$.

\item If $j\neq i+1,i+2$ then $\pi\partial_G(e)=\sum_{x\in S_G\setminus\{a,c,b\}}(e:x)\pi(x)$\\
$=\sum_{x\in S_G\setminus\{a,c,b\}}(e:x)(x\setminus e)=\partial_{G'}(e\setminus c)=\partial_{G'}\pi(e)$.

\end{itemize}

\end{document}